\documentclass{article}
\usepackage[utf8]{inputenc}       
\usepackage[T1]{fontenc}          
\usepackage{graphicx}             
\usepackage{float}                
\usepackage{subfigure}           
\usepackage{caption}             
\usepackage{booktabs}            
\usepackage{multirow}            
\usepackage{longtable}           
\usepackage{tabularx}            
\usepackage{array}               
\usepackage[table]{xcolor}       
\usepackage{amsmath}             
\usepackage{amsfonts}            
\usepackage{nicefrac}            
\usepackage{microtype}           
\usepackage{url}                 
\usepackage{hyperref}            
\usepackage{doi}                 
\usepackage[round]{natbib}       
\usepackage{lipsum}              
\usepackage{verbatim}            
\usepackage{appendix}            
\usepackage{enumitem}            
\usepackage{xcolor}              
\usepackage{listings}            
\usepackage{fancyvrb}            
\usepackage{footmisc}            
\usepackage{algorithm}           
\usepackage{algorithmic}         
\usepackage{authblk}             
\usepackage{ragged2e}            
\usepackage{arxiv}               
\title{VERITAS: Verification and Explanation of Realness in Images for Transparency in AI Systems}

\hypersetup{
pdftitle={A template for the arxiv style},
pdfsubject={q-bio.NC, q-bio.QM},
pdfauthor={David S.~Hippocampus, Elias D.~Striatum},
pdfkeywords={First keyword, Second keyword, More},
}

\author{
Aadi Srivastava\textsuperscript{1,$\dagger$},
Vignesh Natarajkumar\textsuperscript{2,$\dagger$},
Utkarsh Bheemanaboyna\textsuperscript{3},\\
Devisree Akashapu\textsuperscript{2},
Nagraj Gaonkar\textsuperscript{4},
Archit Joshi\textsuperscript{4}
}

\affil{
\textsuperscript{1}Department of Computer Science \& Engineering, Indian Institute of Technology, Madras \\
\textsuperscript{2}Department of Electrical Engineering, Indian Institute of Technology, Madras \\
\textsuperscript{3}Department of Mechanical Engineering, Indian Institute of Technology, Madras \\
\textsuperscript{4}Department of Civil Engineering, Indian Institute of Technology, Madras 
}

\begin{document}
\maketitle

\begingroup
\renewcommand\thefootnote{}
\footnotetext{
\hspace{0pt}
\textsuperscript{$\dagger$} These authors contributed equally to this work. \\
Correspondence to: Aadi Srivastava (aadisrivastava.iitm@gmail.com), Vignesh Natarajkumar (vigneshn@smail.iitm.ac.in)
}
\endgroup

\begin{abstract}
The widespread and rapid adoption of AI-generated content, created by models such as Generative Adversarial Networks (GANs) and Diffusion Models, has revolutionized the digital media landscape by allowing efficient and creative content generation. However, these models also blur the difference between real images and AI-generated synthetic, or “fake”, images, raising concerns regarding content authenticity and integrity. While many existing solutions to detect fake images focus solely on classification and higher-resolution images, they often lack transparency in their decision-making, making it difficult for users to understand why an image is classified as fake. In this paper, we present VERITAS, a comprehensive framework that not only accurately detects whether a small (32x32) image is AI-generated but also explains why it was classified that way through artifact localization and semantic reasoning. VERITAS produces human-readable explanations that describe key artifacts in synthetic images. We show that this architecture offers clear explanations of the basis of zero-shot synthetic image detection tasks. Code and relevant prompts can be found at \url{https://github.com/V-i-g-n-e-s-h-N/VERITAS}
\end{abstract}


\section{Introduction}
The rapid advancement of generative models such as GANs and Diffusion models has democratized access to image generation. Generative Adversarial Networks (GANs) \citep{goodfellowGenerativeAdversarialNetworks2014}, Variational AutoEncoders (VAEs) \citep{kingmaAutoEncodingVariationalBayes2022}, Diffusion models \citep{hoDenoisingDiffusionProbabilistic2020} and other synthetic image generation techniques have created strikingly realistic images and videos. Another important consideration is the growing multimodality of generative models, indicating the ability of models to synthesize audio waveforms and videos along with images \citep{liuSoraReviewBackground2024}, \citep{zhangMMLLMsRecentAdvances2024}, \citep{liM$^3$ITLargeScaleDataset2023}, \citep{maazVideoChatGPTDetailedVideo2024}, \citep{luUnifiedIO2Scaling2023}. As the performance of these generative models progresses, the gap between human-created ‘real’ images and synthetic images shrinks. Naturally, the ability of these models to generate novel samples from training data finds uses in several media-intensive fields. \\

While these innovations empower new possibilities, they also introduce significant challenges. There are risks associated with AI \citep{hendrycksOverviewCatastrophicAI2023}, and AI-generated media is no exception. In fields where transparency and accuracy are very important, the prevalence of AI-generated media creates apprehensions regarding the spread of misinformation and malicious usage of generated media. This raises the concern: \textit{what is real and what is not?}. \\

While several classification systems have emerged to detect synthetic images, they often operate as black boxes, offering limited explainability and seeking to merely improve the accuracy of inference on synthetic data. They also do not always highlight the artifacts used in the classification process. However, to improve trust in and adoption of such frameworks, and to improve fundamental research and understanding in the field of synthetic media detection, such frameworks should be explainable \citep{birdCIFAKEImageClassification2023}, \citep{kangLEGIONLearningGround2025}, \citep{wuExplainableSyntheticImage2025}, \citep{pinhasovXAIBasedDetectionAdversarial2024}, \citep{abirDetectingDeepfakeImages2022}, \citep{malolanExplainableDeepFakeDetection2020}. This will also assist in responding to alterations and adversarial attacks that can be performed on digital media. \\ 

Another consideration is the size of images typically used. Detecting whether small (32x32) images are real or AI-generated is a crucial use case due to the growing prevalence of synthetic media and the unique challenges posed by low-resolution content. Small images are often used as profile pictures, thumbnails, icons, or in messaging apps, making them a common vector for misinformation, identity fraud, and malicious activity. Their limited pixel information makes it harder for both humans and automated systems to distinguish genuine photos from AI-generated fakes, increasing the risk of undetected manipulation. Effective detection in this context helps maintain trust in digital platforms, safeguards user identities, and supports the integrity of online communications.

\subsection{Our Contributions}
We propose VERITAS, a multi-stage framework that classifies small images as either real or synthetic and uses techniques to offer human-readable explanations of the anomalies and artifacts found in synthetic images. Our key contributions are:

\begin{itemize}
    \item Identifying and reviewing several classification frameworks, algorithms, and architectures to accurately detect synthetic images.
    \item Introducing a novel framework that generates artifact-based explanations of the synthetic nature of small images, providing users with clear, concrete evidence for \textit{why} an image is identified as artificial.
    \item Observing the impact of variation in Vision-Language Models (VLMs) and prompting approaches on the produced explanations. 
\end{itemize}

The rest of this paper is structured as follows:
In Section 2 and 3, we review related works in deepfake classification, artifact detection, and explainable AI. In Section 4, we present our methodology, detailing the VERITAS pipeline.
Section 5 describes our experimental setup, including the dataset utilized for evaluations, a description of the reference artifacts, and a discussion of the experimental results. 
Section 6 concludes the paper with an emphasis on the future directions of work identified during the study. Section 7 conveys our acknowledgments to those who supported this study.

\begin{figure}[h!]
    \centering
    \includegraphics[width=0.40\textwidth]{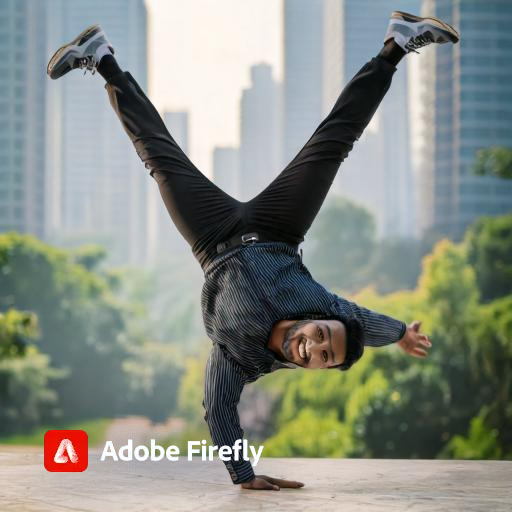}
    \captionsetup{justification=centering}  
    \caption{
        Sample image generated by Adobe Firefly, analyzed by the proposed pipeline.\\
        \textbf{Artifact Description provided by model:} The man’s legs are bent at unnatural angles—his left leg bends backward while the right leg bends forward. The limbs appear disconnected from the torso, resulting in a surreal and anatomically implausible configuration.
    }
    \label{fig:example}
\end{figure}

\section{Related Work}
While the first stage of the proposed pipeline involves robust and accurate classification of real and synthetic images, the primary focus of our innovation is on the explainable analysis of images classified as synthetic. Given that humans cannot always visually discriminate between real and synthetic images, even in the case of images with adversarial attacks, the pipeline is made explainable and understandable by humans. This is in alignment with the goal of Explainable AI (XAI) \citep{gunning_xaiexplainable_2019}: to provide explanations of why a decision was made by a model. \\

The work of \citep{yuAttributingFakeImages2019} highlights that GANs leave “fingerprints” in synthetic images, and minor differences in the generative model can lead to differences in the fingerprints present in an image. Similarly, \citep{geirhosImageNettrainedCNNsAre2019} identify that differences in local texture details in specific patches of the image can lead to different CNN-based classification outcomes. This highlights the importance of local spatial features and artifacts in determining if an image is real or synthetic. These artifacts may be observed via spatial characteristics or frequency-domain features \citep{liIctuOculiExposing2018} \citep{haliassosLipsDontLie2021}, \citep{qianThinkingFrequencyFace2020}, \citep{laiGMDFGeneralizedMultiScenario2024}, \citep{jeongBiHPFBilateralHighPass2021}. \\

It is observed that some local image patches with artifacts contribute more to the classification of the image than others. Hence, patches are selectively chosen and used to generate explanations of artifacts used in the classification of an image as synthetic. To understand which artifacts fake image detectors look for during classification, \citep{chaiWhatMakesFake2020} utilize a fully-convolutional patch-based classifier to focus on local patches of images, and they discovered that using such small receptive fields, such as sliding patches, helps classifiers ignore global differences and focus on local artifacts that are common to several synthetic images. This improves the performance of the pipeline on zero-shot classification with out-of-domain images. To counter the inability of synthetic image detectors to perform well across different generative architectures such as GANs and Diffusion models, \citep{chenSingleSimplePatch2024} propose a Single Simple Patch (SSP) network that extracts noise fingerprint information from a single patch of an image. This approach is supported by \citep{zhongPatchCraftExploringTexture2024}, who observe that texture patches of images reveal more traces of generative models than the global nature of the image. Consequently, global information is erased and texture patches are enhanced for synthetic image detection. These observations motivate our patch-based approach. \\

The detection of artifacts becomes significantly more challenging in the case of the CIFAKE dataset \citep{birdCIFAKEImageClassification2023} (which has been described in greater detail in later sections) owing to the reduced dimensions of the input image and feature map. Further, according to the Information Bottleneck principle \citep{tishbyDeepLearningInformation2015}, there is a possibility of information loss as the feature map is passed through several layers of a network. To counter this, several super-resolution approaches have been explored \citep{anwarDenselyResidualLaplacian2019}, \citep{dongImageSuperResolutionUsing2015}. Several methods proposed earlier can be divided into either interpolation-based approaches \citep{zhangSingleImageSuperResolutionBased2018}, \citep{tongSuperresolutionReconstructionBased2007}, \citep{sunSuperResolutionReconstruction2019} or reconstruction-based approaches \citep{yuSingleimageSuperresolutionBased2017}, \citep{changSingleImageSuper2018}. Previous studies have also used Transformers, such as IPT \citep{chenPreTrainedImageProcessing2021}, SwinIR \citep{liangSwinIRImageRestoration2021}, and the UFormer \citep{wangUformerGeneralUShaped2021}. \citep{hsuDRCTSavingImage2024} propose the Dense-residual-connected Transformer (DRCT) architecture, which utilizes the ability of Swin Transformers \citep{liuSwinTransformerHierarchical2021} to capture long-range dependencies and reconstruct spatial information for preventing such information bottlenecks. Such elements are also included in our pipeline to compensate for information restrictions and downstream losses in layers. \\

Vision-Language Models (VLMs) align textual and visual representations in a shared embedding space, allowing multimodal data processing. Previous studies have established VLMs as powerful tools for multimodal understanding \citep{bommasaniOpportunitiesRisksFoundation2022}, \citep{liuVisualInstructionTuning2023a}, \citep{liuImprovedBaselinesVisual2024}, \citep{chenMiniGPTv2LargeLanguage2023a}, \citep{zhuMINIGPT4ENHANCINGVISIONLANGUAGE}, \citep{qinHowGoodGoogle2023}, \citep{chenVisualGPTDataefficientAdaptation2022}, \citep{liSurveyStateArt2025}. Given the performance of VLMs in complex vision-language tasks such as object localization, fine-grained object description, image-based text generation and detection of abnormalities and defects \citep{qinHowGoodGoogle2023} \citep{chenMiniGPTv2LargeLanguage2023a} \citep{zhuMINIGPT4ENHANCINGVISIONLANGUAGE}, \citep{chenVisualGPTDataefficientAdaptation2022}, integration of these models in frameworks to detect and describe synthetic images is a promising avenue. CLIP \citep{radfordLearningTransferableVisual2021a} is an efficient method of learning image-text relationships and employs an image-text contrastive objective that clusters paired images and texts closer and pushes others away in the embedding space. This allows pre-trained VLMs to capture rich vision-language relationships and is used in our proposed framework for relating patch information with artifacts. CLIP also supports zero-shot predictions based on the developed embedding space, performing very well on the description of unseen images. MOLMO \citep{deitkeMolmoPixMoOpen2024} is a state-of-the-art open-source VLM that outperforms other models such as Qwen \citep{baiQwen25VLTechnicalReport2025} and Pixtral \citep{agrawalPixtral12B2024} on tasks such as fine-grained understanding, pointing and counting of features, localization of artifacts and visual skills. This makes it ideally suited for an artifact-based classification objective. Naturally, VLMs have been used in previous works for synthetic image detection. For instance, \citep{keitaBiLORAVisionLanguageApproach2024} tune VLMs like BLIP-2 to detect synthetic images generated by unseen models using an innovative method called Bi-LORA that combines VLMs with low-rank adaptation tuning. Similarly, \citep{keitaFIDAVLFakeImage2024} uses the complementary relationship between vision and language to detect and assign synthetic images to their source models, prompting their framework to refine outputs. \citep{khanCLIPpingDeceptionAdapting2024} uses Prompt Tuning to adapt CLIP for deepfake detection while retaining both the visual and textual information during inference, improving the accuracy of the classification while also reducing the required training data. \\

Gradient-based methods, such as GradCAM \citep{selvarajuGradCAMVisualExplanations2020}, offer a powerful approach to understanding and explaining the predictions and classifications of CNN-based models, aligning with the principles of Explainable AI (XAI) and interpretability. GradCAM utilizes the gradient of a target concept, such as a classification output, and interprets it along with the feature maps generated by the CNN during training. The result is spatially aligned with the input image, and a visual heatmap that highlights critical regions influencing a model’s predictions is generated. This visual dimension can also be used in the classification process to improve accuracy. Notably, using feature maps without gradients to generalize artifact representations across generative model architectures has been previously explored in \citep{ojhaUniversalFakeImage2024}. \citep{tanLearningGradientsGeneralized2023} uses a pre-trained CNN to convert GAN-generated and real images into GradCAM gradients, which are used as generalized representations of artifacts. These representations can then be leveraged to classify images as real or synthetic, enabling cross-model and cross-dataset robust performance. \citep{birdCIFAKEImageClassification2023} also sought to understand the basis for synthetic image detection using CNNs and discovered that while most regions of a real image are relevant for classification, only a few regions and artifacts of a synthetic image determine its class. This encourages the use of GradCAM to highlight significant patches in images for classification. \\

\section{Review of Classification Frameworks}
Synthetic image classification methods aim to distinguish between real and AI-generated content using binary or multiclass classifiers trained on labeled datasets. Recent advancements in synthetic image detection have focused on leveraging diverse neural architectures and algorithms to distinguish AI-generated images from real ones. This section presents an extensive review of relevant previous research. Classification in the presence of Adversarial perturbations is also considered in the Appendix. 

\subsection{Fundamental Architectures}
Several deepfake classification models primarily rely on Convolutional Neural Networks (CNNs) like ResNet and EfficientNet. For instance, \citep{birdCIFAKEImageClassification2023} train a CNN on the CIFAKE dataset and achieve commendable accuracy by detecting subtle background artifacts rather than semantic object features. \citep{akagicExploringImpactReal2024} investigate model behavior on real and synthetic image inputs and demonstrate that diffusion-generated samples could substitute real images for training classifiers such as CNNs. The evolution of deepfake detection has seen significant contributions from Vision Transformers (ViTs), which address limitations in CNNs regarding global context modeling. Convolutional Neural Networks (CNNs), such as ResNet18 and EfficientNet, demonstrate strong performance in capturing local, low-level features like textures and edge patterns that are often indicative of low-level synthetic artifacts. On the other hand, transformer-based architectures like ViT-Tiny prove effective in identifying global context and long-range dependencies. Recent surveys, such as \citep{wangTimelySurveyVision2024}, highlight that ViT-based models generalize better across deepfake datasets and exhibit robustness under compression and domain shifts. Standalone ViT architectures such as FakeFormer \citep{FakeFormerEfficientVulnerabilityDriven} leverage self-attention to capture long-range dependencies in high-resolution images, though they initially underperformed CNNs due to insufficient local artifact sensitivity. Hybrid models combine ViTs with CNNs or graph networks. For instance, the Self-Supervised Graph Transformer \citep{khormaliSelfSupervisedGraphTransformer2023} integrates a ViT feature extractor with graph convolution layers to improve performance on the FaceForensics++ dataset \citep{rosslerFaceForensicsLearningDetect2019}.\\

\citep{romeoFasterLiesRealtime2024} introduce a Binary Neural Network (BNN) to classify deepfakes and also proposed augmenting the RGB input images with 2 additional filter outputs: the Fast Fourier Transform (FFT) and Local Binary Pattern (LBP). Fast Fourier Transform is employed to include information on frequency-domain patterns created by generative models and frequency-based adversarial attacks as well, which are not discernible by observing RGB images. 
Local Binary Pattern (LBP) complements this by focusing on spatial anomalies. It encodes the intensity differences between a pixel and its neighboring pixels into a binary pattern, capturing local texture irregularities such as texture and artifacts. Together, these filters help identify global and local features of real and synthetic images. A potential bottleneck of this approach in the setting of this paper is the lack of spatial and frequency-domain information that can be reliably extracted from 32x32 images obtained from CIFAKE, leading to irregular results. \citep{emamiSequentialTrainingNeural2023} propose a Sequential Gradient-based Adversarial Training approach to enhance model robustness against progressively sophisticated adversarial attacks. The training begins with a base ResNet50 model \citep{heDeepResidualLearning2015}, initialized with transfer learning weights, and proceeds iteratively. At each stage, adversarial examples are generated using the preceding model, and a new model is trained to resist these attacks. This sequential setup ensures that each subsequent model inherits the weights of its predecessor, with fine-tuning focused primarily on the final layers. Over multiple iterations, the ensemble learns to recognize and counter increasingly complex adversarial attacks. Following the framework of \citep{hintonDistillingKnowledgeNeural2015}, knowledge distillation can be applied by transferring knowledge from a teacher model to a student model to improve model generalization. In our setup, the teacher is a pretrained ResNet18 model, while the student is a similar but lightweight network. The training objective combines the standard cross-entropy loss with a distillation loss that aligns the student’s output distribution with that of the teacher, to help the student learn both the classification outcomes and the feature representations of the teacher. 

\subsection{Ensembling Approaches}
As described earlier, CNNs and Transformers exhibit proficiency in learning local feature maps and modeling long-range relationships, respectively. Previous studies on harnessing the strengths of multiple models in complex classification tasks such as deepfake detection frequently utilize ensembling of models, such as CNNs and Transformer models \citep{zhangVideoDeepfakeClassification2024}, \citep{naskarDeepfakeDetectionUsing2024}, \citep{coccominiCombiningEfficientNetVision2022}, \citep{hoodaD4DetectionAdversarial2022}. Ensembling is a widely adopted strategy in deep learning that leverages the strengths of multiple models to improve classification performance and generalization. Additionally, ensembling provides robustness by reducing the impact of individual model biases and failure cases. These ensembles are often tuned by optimizers to obtain the most optimal relative weighting factors with which outputs of different models in the ensemble are to be combined. \\

Recognizing the complementary strengths of CNN and Transformer architectures, a potential solution for robust classification would be an Optuna-tuned ensemble of pretrained ResNet18, EfficientNet, and ViT-Tiny, which optimizes their contribution weights based on extensive hyperparameter trials, in order to obtain the maximum classification accuracy. The algorithm for such a model is given below

\begin{algorithm}[H]
\caption{Ensemble Model for Image Classification}
\label{alg:ensemble}
\textbf{Input}: Input image $I$ of size $32 \times 32$, training dataset $D$, number of trials $T$ \\
\textbf{Output}: Trained ensemble model and configuration JSON
\begin{algorithmic}[1]
\STATE Initialize models: $M_{\text{ViT}}$, $M_{\text{EfficientNet}}$, $M_{\text{ResNet18}}$
\FORALL{image $i$ in $D$}
    \STATE Extract ViT features: $f_{\text{ViT}} \leftarrow M_{\text{ViT}}(i)$
    \STATE Compute loss: $L_{\text{ViT}} \leftarrow \text{ContrastiveLoss}(f_{\text{ViT}}, \text{labels})$
    \STATE Upsample image: $i_{\text{up}} \leftarrow \text{Upsample}(i, 224\times224)$
    \STATE Extract EfficientNet features: $f_{\text{Eff}} \leftarrow M_{\text{EfficientNet}}(i_{\text{up}})$
    \STATE Compute loss: $L_{\text{Eff}} \leftarrow \text{BCELoss}(f_{\text{Eff}}, \text{labels})$
    \STATE Extract ResNet features: $f_{\text{Res}} \leftarrow M_{\text{ResNet18}}(i_{\text{up}})$
    \STATE Compute loss: $L_{\text{Res}} \leftarrow \text{BCELoss}(f_{\text{Res}}, \text{labels})$
    \STATE Update models using $L_{\text{ViT}}, L_{\text{Eff}}, L_{\text{Res}}$
\ENDFOR
\STATE Initialize Optuna study
\FOR{trial = 1 to $T$}
    \STATE Suggest weights: $w_1, w_2, w_3 \leftarrow \text{trial.suggest\_float}(0, 1)$
    \STATE Normalize weights: $w_i \leftarrow \frac{w_i}{\sum w_i}$
    \STATE Compute ensemble output: $f_{\text{ensemble}} \leftarrow w_1 f_{\text{ViT}} + w_2 f_{\text{Eff}} + w_3 f_{\text{Res}}$
    \STATE Evaluate ensemble performance
    \STATE Update best weights based on validation score
\ENDFOR
\STATE Generate output dictionary with model weights, performance metrics, and model paths
\end{algorithmic}
\end{algorithm}

\subsection{Frequency Domain Analysis}
Apart from the usage of Fast Fourier Transforms (FFT) as described in the earlier subsection, a frequency domain analysis of synthetic images provides a deeper understanding of the signatures left by generative models in the images. Recent advancements in detecting synthetic images have highlighted the importance of examining frequency domain characteristics to differentiate between real and artificial content. For instance, \citep{qianThinkingFrequencyFace2020} propose F3-Net, which uses local frequency statistics for detection. These detection methods involve transforming images from the spatial domain to the frequency domain using techniques such as the Discrete Fourier Transform (DFT) and the Fast Fourier Transform (FFT). \citep{durallUnmaskingDeepFakesSimple2020} and \citep{frankLeveragingFrequencyAnalysis2020} identify abnormal spectral peaks and imbalances in GAN-generated images using frequency analysis as described above, which can be used to identify synthetic images from these models. However, the same frequency artifacts may not be found in images generated from diffusion models, since frequency fingerprints are unique to each generative model architecture. 

\subsection{Directional Enhanced Feature Learning Network}
\citep{liAreHandcraftedFilters2024a} introduce the Direction Enhanced Feature Learning (DEFL) Network, a model using Multi-Directional High-Pass Filters (MHFs) and randomly initialized filters to enhance the diversity of features extracted from input images. 8 well-defined high-pass “base” filters that extract features of the image in 8 different directions are initialized. To improve the diversity of the base filters and to sufficiently capture subtle features, the study further derives a set of 246 "composite” filters according to the 8 base filters. In particular, \( n \leq 7 \) base filters are selected to form a composite filter using the convolution relationship \[
h_c = h_1 \times h_2 \times \dots \times h_n
\]

where \( n \) is the number of base filters chosen.
These together form the set of Multi-Directional High-Pass Filters, 2 of which are randomly repeated to form a total of 256 filters. There are also 256 randomly initialized filters that will be learned during the process of training. The filters are arranged in blocks of 64 MHFs and 64 randomly initialized filters, all of which are learnable and sequentially arranged. The DEFL network produces an output of 256 concatenated feature maps of height and width similar to that of the input image. which can be used for classification, and is rich in extracted features. 

\subsection{Latent Space Optimization}
Given the constraints of working with 32 × 32 images in the CIFAKE and custom datasets, the feature space for extraction is inherently limited. To tackle this, we propose an approach that converts the input images to a high-dimensional space using the embeddings learnt by ViT-Tiny during the training process, following which the boundary between real and synthetic images could be made clearer. Hence, this approach improves the separability of features between real and synthetic images. We utilize the embeddings from the penultimate layer of ViT-Tiny and train them using a combination of Contrastive loss and Triplet loss. The combined loss function is expressed as:  
\begin{equation*}
\mathcal{L} = \alpha \cdot \mathcal{L}_{\text{contrastive}} + \beta \cdot \mathcal{L}_{\text{triplet}},
\end{equation*}  
where $\alpha$ and $\beta$ are weighting coefficients for the respective losses. 
The respective Loss functions are described using the equations below:
\begin{equation*}
\mathcal{L}_{\text{contrastive}} = \frac{1}{N} \sum_{i=1}^{N} \left( y_i \cdot ||f(x_i) - f(x_j)||_2^2 \right.
\left. +  (1 - y_i) \cdot \max\left( 0, m - ||f(x_i) - f(x_j)||_2 \right)^2 \right)  
\end{equation*}
where N is the total number of pairs, $y_i$ is the binary label being 1 for similar pairs, and 0 for dissimilar pairs. $f(x_i)$ is the embedding, and m is the margin of dissimilar pairs.
\begin{equation*}
\mathcal{L}_{\text{triplet}} = \frac{1}{N} \sum_{i=1}^{N} \max \left( 0, ||f(x_i^a) -
f(x_i^p)||_2^2 - ||f(x_i^a) - f(x_i^n)||_2^2) + m \right),
\end{equation*}
\(x_i^a\), \(x_i^p\), and \(x_i^n\) are the anchor, positive, and negative examples respectively. 

Upon visualization using Dimensionality Reduction using t-SNE \citep{JMLR:v9:vandermaaten08a}, while certain features exhibited improved discrimination, others remained challenging to distinguish between real and fake images. The visualization is given below.  
\begin{figure}[h]
    \centering
    \includegraphics[width=0.7\linewidth]{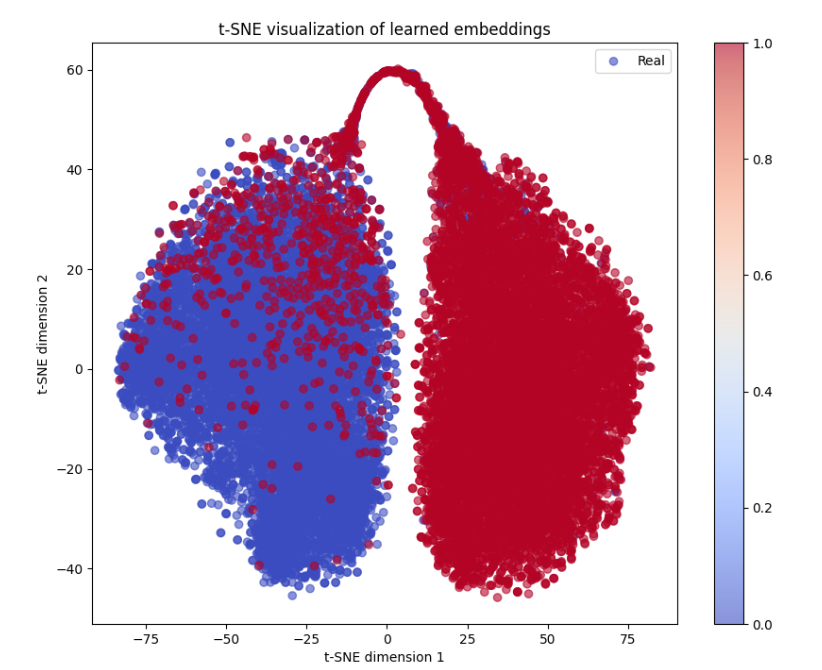}
    \caption{Embeddings trained on contrastive and triplet loss}
    \label{fig:enter-label}
\end{figure}

To evaluate the utility of these modified embeddings, they can be passed directly to the ViT-Tiny model, bypassing its initial input layer. Further improvements in performance can be obtained with appropriate optimization of the weighting coefficients for the Contrastive and the Triplet Losses, an approach that has not been explored in the current study. 

\begin{algorithm}[H]
\caption{Contrastive Loss Computation}
\label{alg:contrastive}
\textbf{Input}: Feature matrix $F$, label vector $y$, temperature $\tau$ \\
\textbf{Output}: Contrastive loss $L_{\text{pos}}$
\begin{algorithmic}[1]
\STATE Normalize features: $F \leftarrow \text{normalize}(F)$
\STATE Compute similarity matrix: $S \leftarrow \frac{F F^\top}{\tau}$
\STATE Create masks: $M^+ \leftarrow (y = y^\top)$, $M^- \leftarrow (y \neq y^\top)$
\STATE Remove self-similarity: $M^+ \leftarrow M^+ - I$
\STATE Compute exponentials: $P \leftarrow \exp(S)$
\STATE Compute log probabilities: $\text{log\_prob} \leftarrow S - \log(\sum_{j} \exp(S_{ij}))$
\STATE Compute positive loss: $L_{\text{pos}} \leftarrow -\frac{\sum(M^+ \odot \text{log\_prob})}{\sum M^+}$
\STATE \textbf{return} $L_{\text{pos}}$
\end{algorithmic}
\end{algorithm}

\subsection{Dual-Domain Strategies}
\citep{yanD2DefendDualDomainBased2021} propose a dual-domain filtering approach that combines spatial-domain processing with transform-domain analysis to remove adversarial noise. The features of the architecture are:
\begin{itemize}
    \item \textbf{Spatial Domain Filtering} -  It involves separating the input image into edge (high-frequency) and texture (low-frequency) features. Edge and Texture Decomposition is performed by Bilateral Filtering to decompose the spatial features into 2 components- Filtered (which are rich in high-frequency structural information) and Unfiltered texture features (which contain smoother variations and finer details).
    \item \textbf{Transform Domain Filtering} - The unfiltered texture features are converted into the frequency domain using a Short-Time Fourier Transform (STFT). This facilitates localized frequency analysis, targeting perturbations in specific frequency bands. Wavelet shrinkage is applied in the frequency domain to selectively attenuate high-frequency components associated with adversarial perturbations.
    \item \textbf{Reconstruction} - After the dual-domain filtering, the refined texture features are reconstructed and combined with the preserved edge features to produce a denoised and robust image.
\end{itemize}
 

\subsection{AutoEncoder Architectures}
AutoEncoders have been extensively used in deepfake and adversarial image classification pipelines for two primary purposes: feature purification and anomaly detection, both in the context of adversarial perturbations. The central idea is that AutoEncoders, when trained only on real images, struggle to accurately reconstruct synthetic or adversarially perturbed images. This property can be exploited to flag anomalous reconstructions as potential fakes. One proposed direction involves training the AutoEncoder to reconstruct perturbation maps from adversarial samples and learning to isolate the added noise and distortions. Perturbation maps illustrate the alterations made to adversarial images in comparison to their original counterparts. This purification process allows any following classifiers to focus on underlying semantic features rather than being misled by adversarial artifacts. Previous studies support this approach and the above observations \citep{kalariaAdversarialPurificationUsing2022}, \citep{hwangPuVAEVariationalAutoencoder2019}, \citep{beggelRobustAnomalyDetection2019}. \citep{guoCounteringAdversarialImages2018} proposed Variational AutoEncoder-based purification methods that map adversarial examples to the distribution of clean images before reclassification. \citep{duGeneralizableDeepfakeDetection2020} proposes a Locality-Aware AutoEncoder that learns intrinsic features of images instead of superficial characteristics of training data, enabling better performance on zero-shot classification of deepfakes.

\section{Proposed Framework}
This section presents our core contribution: A five-stage pipeline designed for detecting, interpreting, and describing artifacts in low-resolution images. Our pipeline integrates super-resolution, attention-based heatmapping, patch-level analysis, vision-language modeling, and multimodal language generation to provide comprehensive artifact detection with interpretability.

The five steps are:
\begin{enumerate}
\item Super-resolve input image using DRCT
\item Localize relevant regions with GradCAM heatmaps
\item Divide image into patches weighted by attention
\item Use CLIP-based similarity for artifact classification
\item Generate textual explanations via MOLMO
\end{enumerate}

\begin{figure}[H]
    \centering
    \includegraphics[width=1.05\textwidth]{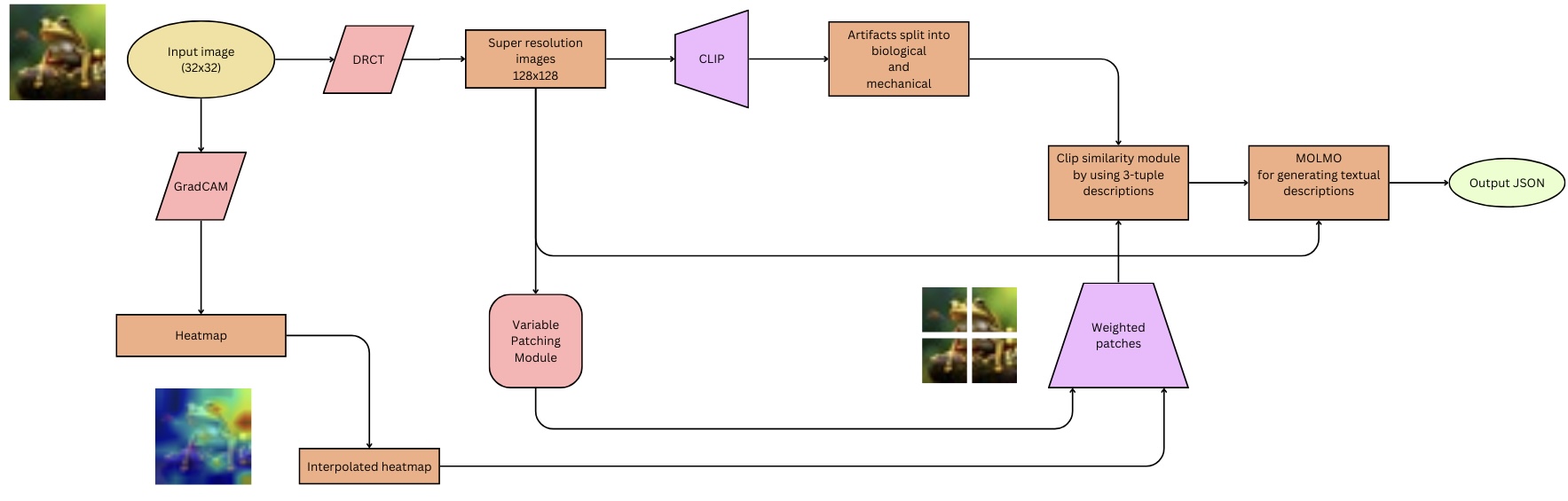} 
    \caption{Pipeline for Detecting Artifacts Using Super-Resolution and GradCAM}
    \label{fig:pipeline}
\end{figure}

\subsection{Step 1: Super-Resolution of Images Using DRCT}
Detecting fine-grained artifacts in low-resolution images, such as those sized 32×32, is inherently difficult due to the lack of detailed spatial information. While upscaling such images using standard interpolation methods such as Bilinear, Bicubic, and Lanczos Interpolation \citep{khaledyan_low-cost_2020}, \citep{bituin_ensemble_2024} can improve visibility, these techniques often introduce artificial smoothing or ringing effects that distort the original image structure. These distortions can obscure or mimic true artifacts, making reliable analysis more difficult.\\

To address this issue, we employ deep learning-based super-resolution methods, which reconstruct high-resolution images by learning mappings from low-resolution inputs to high-resolution outputs. In our work, we utilize DRCT (Dense-Residual-Connected Transformer), a recent hierarchical super-resolution model proposed by Hsu et al.\citep{hsuDRCTSavingImage2024}, which
utilizes the ability of Swin Transformers \citep{liuSwinTransformerHierarchical2021} to capture long-range dependencies and reconstruct spatial
information for preventing information bottlenecks. This architecture allows DRCT to maintain high-frequency visual features and refine textures and edges more effectively than traditional recursive or interpolative approaches. 

\subsection*{Step 2: GradCAM Heatmap Generation}
To interpret the decisions made by our binary classifier in our classification pipeline, we employ GradCAM (Gradient-weighted Class Activation Mapping) \citep{selvarajuGradCAMVisualExplanations2020}, a technique that highlights the regions of an image most influential in the model’s prediction. Specifically, GradCAM helps us visualize which areas the model focuses on when classifying an image as Real or Fake. \\

The method computes the gradient of the target class score with respect to the feature maps from the final convolutional layer. These gradients are then used to weight the feature maps, generating a class-specific activation map that highlights the most salient regions contributing to the model’s decision. By applying a ReLU operation, only positive influences are retained, resulting in a heatmap that overlays the input image. This heatmap serves two primary functions:

This heatmap serves two key purposes:
\begin{itemize}
    \item Interpretability – It provides a visual explanation of the model’s classification by indicating which image regions were most influential.
    \item Artifact localization – It assists in detecting potential artifacts or unnatural features in fake images by highlighting anomalous image areas.
\end{itemize}

Importantly, we apply GradCAM on the original 32$\times$32 images, not on resized or super-resolved versions. This ensures that the attention corresponds directly to the image used during classification, avoiding the risk of introducing analysis distortions from post-processing. Despite the low resolution, GradCAM still provides meaningful spatial cues, supporting artifact detection.

\subsection*{Step 3: Patching and Weighing Based on GradCAM Heatmap}
For images passed through DRCT, where artifacts are influenced or revealed more distinctly through super-resolution, we perform spatial patch-based analysis to localize and quantify regions that are likely to contain visual anomalies. After applying super-resolution, the resulting high-resolution image is divided into variable-sized patches to enable fine-grained analysis of local regions.\\

To determine the importance of each patch, we utilize the GradCAM heatmap computed in Step 2. Since this heatmap was originally generated on the low-resolution (32$\times$32) input image, we interpolate it to match the spatial dimensions of the super-resolved image. Each patch is then assigned a weight that reflects the importance of that patch for artifact detection as indicated by the heatmap. The weight $w_{\text{patch}}$ of a given patch is computed as the summation of heatmap intensities over all pixels within the patch:

\begin{equation}
w_{\text{patch}} = \sum_{i \in \text{patch}} H(i),
\end{equation}

where $H(i)$ denotes the GradCAM heatmap intensity at pixel $i$.

\subsection*{Step 4: CLIP-Based Probability Score for Artifacts}
To simplify and improve the process of identifying artifacts, we employ a CLIP-based classification strategy \citep{radfordLearningTransferableVisual2021a}, since CLIP is an efficient method of learning image-text relationships. Initially, we apply a binary classifier using CLIP similarity to distinguish between two broad semantic categories: animals and vehicles. These categories were chosen due to their highly separable embeddings in the CLIP latent space, which allows us to obtain a well-defined decision boundary.

For more targeted artifact detection, we curated class-specific artifact descriptors for each category. Each artifact is described using a three-tuple format:
\begin{itemize}
    \item \textbf{Positive Description}: a description indicating that the patch contains the artifact being described. 
    \item \textbf{Negative Description}: a description indicating that the patch does not contain the artifact being described, and is realistic. 
    \item \textbf{Neutral Description}: a description indicating that the patch layout is not relevant to the artifact being searched for. For instance, an artifact relating to edge layouts cannot be present in a patch that has no edges. 
\end{itemize}

The descriptions were constructed keeping in mind that the CLIP score is affected by semantic similarity.\\

For each patch (as defined in Step 3), we compute CLIP similarity scores against the three descriptors. Based on the highest similarity, each patch casts a vote as \textbf{positive}, \textbf{negative}, or \textbf{neutral}. These votes are then aggregated using a weighted average based on patch importance derived from the GradCAM heatmap. The overall artifact score $S$ is computed using:

\begin{equation}
S = \frac{\sum_k w_k v_k}{\sum_k w_k},
\end{equation}

where $w_k$ is the weight of patch $k$ and $v_k$ is its corresponding vote (numerically encoded). A higher value of $S$ indicates stronger evidence of artifacts within the image. Patches (and thus images) that exceed a predefined threshold are retained for further inspection, and the others are filtered out.

\subsection*{Step 5: Generating Textual Descriptions Using MOLMO}
To enhance artifact detection with interpretable, human-readable insights, we integrate MOLMO \citep{deitkeMolmoPixMoOpen2024}, a state-of-the-art open-source Vision-Language Model (VLM). For images containing patches identified as likely to exhibit artifacts, MOLMO is prompted to generate detailed textual descriptions that articulate the type and characteristics of the detected artifacts. This step enables the system to produce explanations that are both contextually relevant and semantically informative, thereby improving the interpretability of the analysis.

The algorithmic flow of the proposed framework is given below. 
\begin{algorithm}[H]
\caption{Artifact Detection and Explanation Pipeline}
\label{alg:artifact_pipeline}
\textbf{Input}: Low-resolution image $I_{lr}$ (e.g., 32$\times$32) \\
\textbf{Output}: Artifact detection score $S$ and textual explanation $E$
\begin{algorithmic}[1]
\STATE \textbf{Step 1: Super-Resolution}
    \STATE Super-resolve $I_{lr}$ using DRCT to obtain high-resolution image $I_{sr}$
\STATE \textbf{Step 2: GradCAM Heatmap Generation}
    \STATE Apply binary classifier on $I_{lr}$ to obtain class prediction
    \STATE Generate GradCAM heatmap $H$ on $I_{lr}$
\STATE \textbf{Step 3: Patching and Weighting}
    \STATE Interpolate $H$ to match dimensions of $I_{sr}$
    \STATE Divide $I_{sr}$ into patches $\{p_k\}$
    \FORALL{patch $p_k$}
        \STATE Compute patch weight $w_k$ as sum of heatmap intensities in $p_k$
    \ENDFOR
\STATE \textbf{Step 4: CLIP-Based Artifact Classification}
    \FORALL{patch $p_k$}
        \STATE Compute CLIP similarity with positive, negative, and neutral descriptors
        \STATE Assign vote $v_k$ based on highest similarity descriptor
    \ENDFOR
    \STATE Compute artifact score: $S = \frac{\sum_k w_k v_k}{\sum_k w_k}$
    \STATE Filter images with $S$ below threshold
\STATE \textbf{Step 5: Generating Textual Explanation}
    \IF{$S$ exceeds threshold}
        \STATE Generate textual explanation $E$ using MOLMO for detected artifacts
    \ENDIF
\end{algorithmic}
\end{algorithm}
\section{Experimental Results}
\subsection{Evaluation dataset}
As generative models improve, distinguishing real from synthetic media becomes increasingly challenging. CIFAKE \cite{birdCIFAKEImageClassification2023} directly addresses this issue by providing a balanced and labeled open-source dataset for binary classification. CIFAKE is generated using Stable
Diffusion 1.4 (SDM), an example of Latent Diffusion Models used to generate images \citep{rombachHighResolutionImageSynthesis2022}. The SDM generates a synthetic equivalent of CIFAR-10, a well-known dataset of natural images across ten classes. Each image in CIFAKE is 32×32 pixels in resolution. The dataset consists of 120,000 labeled images, evenly split between real and synthetic samples. The real images are sourced from the CIFAR-10 dataset, while the fake counterparts are generated using the Stable Diffusion 1.4 model.\\

Unlike many deepfake or forgery detection datasets, such as FaceForensics++\citep{rosslerFaceForensicsLearningDetect2019}, which provide higher-resolution images, CIFAKE generalizes the detection task to broader categories at a much lower resolution. This introduces unique challenges: low-resolution images inherently contain less semantic and structural information, making it more difficult to detect subtle generative artifacts. Many existing detection architectures often require upsampling or significant architectural modification to function effectively on CIFAR-sized inputs. Smaller images also pose the challenges of identifying whether the distortion is due to the small resolution or is an artifact introduced due to AI generation.

\subsection{A Comparison of VLMs}
To identify the right Vision-Language Model for our use case, we perform a comparative study of 3 popular VLMs with the same prompt, and evaluate the quality of the responses. These 3 models are:
\begin{itemize}
    \item \textbf{MOLMO} \citep{deitkeMolmoPixMoOpen2024} is an open-source family of state-of-the-art vision-language models (VLMs) developed by the Allen Institute for AI. Trained on the highly curated PixMo dataset, Molmo excels at image understanding, pointing, and object localization, outperforming many larger proprietary models while remaining fully open and reproducible
    \item \textbf{Qwen 2.5 VL} \citep{baiQwen25VLTechnicalReport2025} is the latest vision-language model in the Qwen series by Alibaba Group, offering advanced visual recognition, document parsing, and long-video understanding. 
    \item \textbf{Pixtral 12B} \citep{agrawalPixtral12B2024} is Pixtral 12B AI’s flagship 12-billion-parameter VLM, designed for robust performance across both text and multimodal tasks. It features a novel architecture with a 400M vision encoder and excels in document comprehension, chart analysis, and instruction following
\end{itemize}

A collection of fake images from CIFAKE is sampled, and the characteristic artifacts for these images are determined in advance. A sample of these images is given below. 

\begin{figure}[H]
\centering
\begin{tabular}{cccc}
\includegraphics[width=3.2cm]{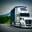} &
\includegraphics[width=3.2cm]{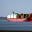} &
\includegraphics[width=3.2cm]{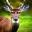} &
\includegraphics[width=3.2cm]{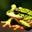} \\
(a) Misaligned Body Panels & (b) Non-manifold Geometries & (c) Biological Asymmetry & (d) Impossible Joints \\
\end{tabular}
\begin{tabular}{cc}
\includegraphics[width=3.2cm]{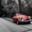} &
\includegraphics[width=3.2cm]{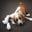} \\
(e) Car Panel Misalignment & (f) Incorrect Paw Structures \\
\end{tabular}
\caption*{Images corresponding to artifact types listed in Table~\ref{tab:vlm_comparison}.}
\end{figure}
 With this ground truth of artifacts in mind, the 3 models are compared for accuracy and descriptiveness of their evaluations. A sample of results on the above images are described below.
\begin{table}[H]
\centering
\small
\renewcommand{\arraystretch}{1.8}
\begin{tabular}{|p{3cm}|p{3.8cm}|p{3.8cm}|p{3.8cm}|}
\hline
\textbf{Artifact Category} & \textbf{MOLMO} & \textbf{Qwen2.5 VL} & \textbf{Pixtral 12B} \\
\hline
Misaligned Body Panels & 
The truck’s body panels are not properly aligned, creating an uneven and distorted appearance. & 
Slight misalignment, particularly noticeable at the panel edges. & 
Misalignment around the front grille and lower edge of the cab. \\
\hline
Non-manifold Geometries in Rigid Structures & 
The ship’s hull shows sharp, unrealistic edges disrupting the solid form. & 
Non-manifold geometries at sharp edges and corners. & 
Non-manifold geometry at cargo container intersections. \\
\hline
Biological Asymmetry Errors & 
The deer’s eyes are asymmetrical, with one appearing larger than the other. & 
Antlers show noticeable size and shape differences. & 
One antler is more developed than the other, creating asymmetry. \\
\hline
Anatomically Impossible Joint Configurations & 
Frog joints and limbs are twisted in anatomically impossible ways. & 
Unnatural leg posture that defies natural joint configuration. & 
Legs are bent in ways not physically possible for real frogs. \\
\hline
Misaligned Body Panels & 
Visible misalignment near the left front wheel arch. & 
Misalignment near the trunk lid causing visible gaps. & 
Misalignment near the rear left wheel arch with visible gaps. \\
\hline
Anatomically Incorrect Paw Structures & 
Dog’s paw has fused digits and an extra digit on the other paw. & 
Twisted and fused digits in the front left paw. & 
Extra fused digit causing asymmetrical paw appearance. \\
\hline
\end{tabular}
\vspace{2pt}
\caption{Comparison of VLM (Vision-Language Model) responses to different artifact prompts across models: MOLMO, Qwen2.5 VL, and Pixtral 12B.}
\label{tab:vlm_comparison}
\end{table}
\vspace*{-1em}
MOLMO has 4 variants: MOLMO-72B, MOLMO-7B-D, MOLMO-7B-O, and MOLMOE-1B. For all our pipeline-based evaluations, we used MOLMO-7B-D to have a moderate dependence on the size of the model.

The following is the general summary of the VLM models:
\begin{itemize}
    \item MOLMO: Descriptions are visually expressive, and the model tends to highlight visible distortions directly. It is particularly effective for generating high-level summaries and professional artifact reviews.
    \item Qwen2.5 VL: This model leverages technical and biological terminology, often producing more precise and domain-specific outputs. It is especially suitable for scientific and clinical interpretation tasks.
    \item Pixtral 12B: The generated descriptions are straightforward and easily readable, making this model well-suited for general audiences or educational purposes. However, it may sometimes lack the nuance or specificity offered by the other models.
\end{itemize}
Based on this comparative evaluation, we select MOLMO for our analysis due to its visually expressive and comprehensive descriptions, which effectively highlight visible distortions and support high-level artifact interpretation.

\subsection{Localization Properties of MOLMO}
MOLMO is particularly well-suited for artifact localization tasks due to its attention-guided vision-language alignment mechanism, which enables the model to focus on specific regions within an image during caption generation. Unlike general-purpose Vision-Language Models such as Qwen2.5 VL or Pixtral 12B, which tend to produce global or loosely grounded descriptions, MOLMO leverages cross-modal attention maps to align linguistic tokens with spatially relevant visual features. This design allows MOLMO to produce fine-grained descriptions such as “The deer’s eyes are asymmetrical, with one appearing larger than the other” instead of generic outputs like “Antlers show noticeable size and shape differences.” \citep{hannan_foundation_2025}

Therefore, MOLMO’s localization-aware architecture provides a distinct advantage in detecting and describing not only what artifact is present in an image, but also where it occurs, making it an ideal choice for fine-grained artifact analysis.

\subsection{Performance of the proposed pipeline}
The following example demonstrates the descriptive capabilities of our proposed pipeline. Among the many artifacts identified across the dataset, a selected subset is presented below to compare the descriptions generated by MOLMO and those produced by our method. 

\begin{longtable}{>{\raggedright\arraybackslash}p{0.45\textwidth} >{\raggedright\arraybackslash}p{0.45\textwidth}}
\caption{Artifact List – Proposed Pipeline vs MOLMO} \\
\toprule
\textbf{Proposed Pipeline} & \textbf{MOLMO} \\
\midrule
\endfirsthead

\toprule
\textbf{Artifact Name (Proposed Pipeline)} & \textbf{Description by MOLMO} \\
\midrule
\endhead

Abruptly cut off objects & Inconsistent object boundaries \\
Ghosting effects: Semi-transparent duplicates of elements & Discontinuous surfaces \\
Dental anomalies in mammals & Floating or disconnected components \\
Anatomically incorrect paw structures & Asymmetric features in naturally symmetric objects \\
Unrealistic eye reflections & Misaligned bilateral elements in animal faces \\
Misshapen ears or appendages & Irregular proportions in mechanical components \\
Unnatural pose artifacts & Texture bleeding between adjacent regions \\
Biological asymmetry errors & Texture repetition patterns \\
Impossible foreshortening in animal bodies & Over-smoothing of natural textures \\
Anatomically impossible joint configurations & Artificial noise patterns in uniform surfaces \\
Asymmetric features in naturally symmetric objects & Unrealistic specular highlights \\
Misaligned bilateral elements in animal faces & Inconsistent material properties \\
Incorrect perspective rendering & Dental anomalies in mammals \\
Scale inconsistencies within single objects & Anatomically incorrect paw structures \\
Spatial relationship errors & Improper fur direction flows \\
Scale inconsistencies within the same object class & Unrealistic eye reflections \\
Depth perception anomalies & Misshapen ears or appendages \\
Improper fur direction flows & Physically impossible structural elements \\
Inconsistent object boundaries & Inconsistent scale of mechanical parts \\
Blurred boundaries in fine details & Incorrect perspective rendering \\
Over-sharpening artifacts & Scale inconsistencies within single objects \\
Excessive sharpness in certain image regions & Spatial relationship errors \\
Aliasing along high-contrast edges & Depth perception anomalies \\
Jagged edges in curved structures & Over-sharpening artifacts \\
Fake depth of field & Aliasing along high-contrast edges \\
Artificial depth of field in object presentation & Blurred boundaries in fine details \\
Discontinuous surfaces & Loss of fine detail in complex structures \\
Texture bleeding between adjacent regions & Artificial smoothness \\
Texture repetition patterns & Movie-poster like composition of ordinary scenes \\
Over-smoothing of natural textures & Dramatic lighting that defies natural physics \\
Regular grid-like artifacts in textures & Artificial depth of field in object presentation \\
Artificial noise patterns in uniform surfaces & Unnaturally glossy surfaces \\
Random noise patterns in detailed areas & Synthetic material appearance \\
Repeated element patterns & Multiple inconsistent shadow sources \\
Systematic color distribution anomalies & Exaggerated characteristic features \\
Unnatural color transitions & Impossible foreshortening in animal bodies \\
Color coherence breaks & Scale inconsistencies within the same object class \\
Frequency domain signatures &  \\
Artificial smoothness &  \\
Exaggerated characteristic features &  \\
Synthetic material appearance &  \\
Inconsistent material properties &  \\
Loss of fine detail in complex structures &  \\
Resolution inconsistencies within regions &  \\
Unrealistic specular highlights &  \\
\bottomrule
\end{longtable}

\newpage
Clearly, our proposed approach is more effective than MOLMO in detecting and helping users understand spatial and frequency-based artifacts, especially in resource-constrained image dimensions. The following examples demonstrate the performance of our pipeline for a few images and selected artifacts. 

\begin{figure}[H]
    \centering
    \subfigure[Example 1]{
        \includegraphics[width=3.15cm]{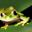}
        \label{fig:img1}
    }
    \subfigure[Example 2]{
        \includegraphics[width=3.15cm]{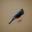}
        \label{fig:img2}
    }
    \caption{Examples 1 and 2.}
    \label{fig:two_examples}
\end{figure}

\begin{table}[H]
\centering
\small
\caption{Descriptions by Proposed Pipeline vs MOLMO for a few artifacts (Example 1)}
\begin{tabular}{>{\raggedright\arraybackslash}p{0.20\textwidth} >{\raggedright\arraybackslash}p{0.37\textwidth} >{\raggedright\arraybackslash}p{0.37\textwidth}}
\toprule
\textbf{Artifact Name} & \textbf{Description by Proposed Pipeline} & \textbf{Description by MOLMO} \\
\midrule

Abruptly cut-off objects & Visible cut-off edge of an object, possibly a car or vehicle, with a noticeable gap between the main body and a protruding element. & Objects that appear to be suddenly and artificially truncated, disrupting the natural composition.\\

Ghosting effects: Semi-transparent duplicates of elements & Visible ghosting effects near the center of the image, showing semi-transparent duplicates of the main elements, creating an illusion of double vision or afterimages. & Semitransparent duplicates of elements, creating a ghostly or ethereal appearance.\\

Anatomically incorrect paw structures & Clear misalignment of frog's front legs, with left leg appearing much larger and distorted compared to the right. This anatomical discrepancy is visible near the frog's body, creating an unnatural and unrealistic appearance. & Paws that don’t accurately represent the natural anatomy of the animal.\\

Unrealistic eye reflections & Clear misalignment of eye reflections, with the right eye appearing significantly lighter than the left. This discrepancy in shading and highlights creates an unrealistic and visually jarring effect on the frog's face. & Artificial or incorrect reflections in eyes, creating an unnatural look.\\

\bottomrule
\end{tabular}
\end{table}

\begin{table}[H]
\centering
\small
\caption{Descriptions by Proposed Pipeline vs MOLMO for a few artifacts (Example 2)}
\begin{tabular}{>{\raggedright\arraybackslash}p{0.20\textwidth} >{\raggedright\arraybackslash}p{0.37\textwidth} >{\raggedright\arraybackslash}p{0.37\textwidth}}
\toprule
\textbf{Artifact Name} & \textbf{Description by Proposed Pipeline} & \textbf{Description by MOLMO} \\
\midrule

Artificial smoothness & Visible artificial smoothness artifact in the image, likely caused by a flaw in the image capture process or post-processing technique. This smooth area contrasts with the surrounding texture, creating an unusual visual effect. & Excessive uniformity; Unnatural, overly polished surface lacking texture and detail.
\\

Improper fur direction flows & Clear artifact visible near the left front wheel arch. Misaligned fur direction flows diagonally from top-right to bottom-left, creating an unnatural appearance and disrupting the vehicle's sleek profile. & Unnatural flow of fur or hair-like features, inconsistent with natural behavior.
\\

Texture repetition patterns & Visible texture repetition patterns in the image, likely indicating a manufacturing defect or material anomaly in the vehicle's surface. & Repetitive motifs in the object's surface, lacking the complexity of real-world textures.
\\

Frequency domain signatures & Visible frequency domain signature artifact near the center of the image, slightly to the left. It appears as a distorted, blurred area with a checkered pattern effect, likely caused by a flaw in the image acquisition or processing. & Distinct patterns in the image's frequency spectrum, often indicating artificial generation or heavy processing.\\

\bottomrule
\end{tabular}
\end{table}

The following examples further demonstrate the usefulness of the pipeline, with 3 artifacts sampled instead of 4, from the numerous artifacts identified by our pipeline and MOLMO. 

\begin{figure}[H]
    \centering
    \subfigure[Example 3]{
        \includegraphics[width=3.15cm]{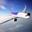}
        \label{fig:img2}
    }
    \subfigure[Example 4]{
        \includegraphics[width=3.15cm]{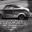}
        \label{fig:img4}
    }
    \caption{Examples 3 and 4.}
    \label{fig:two_examples}
\end{figure}

\begin{table}[ht]
\centering
\small
\caption{Descriptions by Proposed Pipeline vs MOLMO for a few artifacts (Example 3)}
\begin{tabular}{>{\raggedright\arraybackslash}p{0.20\textwidth} >{\raggedright\arraybackslash}p{0.37\textwidth} >{\raggedright\arraybackslash}p{0.37\textwidth}}
\toprule
\textbf{Artifact Name} & \textbf{Description by Proposed Pipeline} & \textbf{Description by MOLMO} \\
\midrule

Incorrect reflection mapping & Noticeable mismatch in reflection between the top and bottom sections of the vehicle's surface, creating an unrealistic or inconsistent appearance. & Mismatched light reflections creating an unrealistic appearance on the aircraft's surfaces\\

Impossible mechanical joints & Clear misalignment of mechanical components near the vehicle's center, creating an unnatural gap between parts. This impossible joint appears to defy physics, with misaligned edges and an irregular space between the mechanical elements. & Joints that defy physics and engineering principles, appearing impossibly structured or positioned\\

Unnaturally glossy surfaces & Visible in the top right corner, extending towards the center. Appears as unnaturally smooth and reflective areas contrasting with the surrounding matte surfaces. These glossy patches are noticeable on the vehicle's exterior, likely indicating a manufacturing defect or intentional design element that stands out from the rest of the body. & Excessive shine on certain areas of the plane, giving it an artificial and unrealistic look\\

\\
\bottomrule
\end{tabular}
\end{table}

\begin{table}[H]
\centering
\small
\caption{Descriptions by Proposed Pipeline vs MOLMO for a few artifacts (Example 4)}
\begin{tabular}{>{\raggedright\arraybackslash}p{0.20\textwidth} >{\raggedright\arraybackslash}p{0.37\textwidth} >{\raggedright\arraybackslash}p{0.37\textwidth}}
\toprule
\textbf{Artifact Name} & \textbf{Description by Proposed Pipeline} & \textbf{Description by MOLMO} \\
\midrule

Floating or disconnected components & Visible disconnect between vehicle body and wheel well, creating an illusion of floating or disconnection near the left front wheel arch. & Front bumper appears to be disconnected from the car body.

\\

Unnaturally glossy surfaces & Visible on the vehicle's exterior, particularly on the door panels and roof. Excessively reflective areas contrast with the matte black paint, creating an unnatural appearance. This glossy effect is noticeable on the left side of the vehicle, where it stands out against the intended matte finish. & The car's surface has an excessive shine that doesn't match real-world lighting conditions\\


Unrealistic specular highlights & Noticeable unrealistic specular highlights visible on the vehicle's surface, particularly on the front right quarter panel and side panels. These highlights appear inconsistent with the vehicle's surroundings and create an artificial, computer-generated look. & There are overly bright, artificial highlights on the car's surface\\

\bottomrule
\end{tabular}
\end{table}
\vspace{1cm}
From the above experiments, it is clear that the proposed pipeline offers the following advantages over conventional synthetic image detection frameworks
\begin{itemize}
    \item Deeper and more fine-grained artifact-based analysis of synthetic images
    \item Clear explanation of realness in small images of sizes 32x32
    \item Highly understandable explanations of why an image is considered synthetic. 
\end{itemize}

\section{Conclusion and Future Work}
The proposed framework works by combining three powerful techniques to detect and explain visual artifacts in generated images. First, it uses a specialized image upscaling method that focuses on preserving fine details like edges and textures, which are often where subtle artifacts appear. Next, the image is broken into patches and analyzed using attention maps (like Grad-CAM) to focus on the regions most relevant to the model’s classification decision. Finally, for the most important patches, a Vision-Language Model generates human-readable descriptions that explain what kind of artifact is present and where it appears. By linking low-level visual signals with high-level semantic descriptions, this framework provides both precise localization and meaningful interpretation of artifacts, making it a powerful tool for analyzing manipulated or AI-generated images.

While the proposed Task 2 pipeline demonstrates a structured and interpretable approach to artifact detection and explanation, several areas offer opportunities for further refinement:

\subsection{Aggregation of GradCAM Attention Maps}
The current system relies on a single-model GradCAM heatmap to localize discriminative regions. However, GradCAM is inherently model-dependent, and its saliency maps may reflect biases unique to a specific network’s learned representations. Aggregating GradCAM maps from multiple classifiers could produce a consensus-based attention signal, reinforcing regions consistently identified across models. This ensemble-based strategy may reduce model-specific artifacts and improve the robustness of localization.

\subsection{Mitigation of Language Model Hallucinations}
Vision-language models like MOLMO can generate fluent but factually unsupported descriptions, introducing hallucinations that undermine interpretability. To address this, future work will incorporate hallucination correction techniques such as Woodpecker\citep{yinWoodpeckerHallucinationCorrection2024}, designed to filter or re-rank model outputs based on factual consistency. Such methods can improve the reliability of textual explanations and prevent misleading interpretations.

\subsection{Improved Image Resizing Strategies}
Interpolation-based upscaling methods may introduce artificial artifacts that mimic or mask real anomalies. Future iterations of the pipeline could adopt more advanced super-resolution techniques, such as edge-preserving or perceptually-aware models, that retain structural fidelity while minimizing introduced distortions. This would further distinguish true generative artifacts from resizing artifacts.

\subsection{Adversarial Robustness with Certified Defenses}
The current system does not address adversarial perturbations, which can distort classification outcomes and mislead GradCAM-based localization. Such perturbations may compromise the interpretability of the pipeline by altering attention maps or affecting semantic scoring. To address this, certified robustness techniques like Auto-LiRPA\citep{xuAutomaticPerturbationAnalysis2020} can be employed. Auto-LiRPA computes provable activation bounds under input perturbations, enabling the training of models that maintain stable predictions within a defined epsilon-neighborhood. Integrating this into our pipeline would enhance resilience and preserve interpretability even under adversarial conditions

\subsection{Generalization to Diverse Generative Artifacts}
Different generative models, such as diffusion models (e.g., Stable Diffusion), transformers (e.g., PixArt), and autoencoders, produce artifacts with distinct visual characteristics. Diffusion models may generate overly smooth textures, while autoencoders often introduce repetitive patterns or edge degradation. These differences arise from architectural and training variations, leading to unique “fingerprints” in generated images.

If a classifier is trained on artifacts from a limited set of models, it risks overfitting to those model-specific cues, reducing its ability to generalize to unseen or emerging generators. To address this, future work should explore domain generalization methods that encourage the model to learn invariant features across multiple generative domains. Additionally, prompt-based adaptation, especially with models like CLIP~\citep{radfordLearningTransferableVisual2021a}, can help tailor detection strategies to diverse image types. These approaches would improve robustness and adaptability as generative models continue to evolve.

\section{Acknowledgment}
The authors thank Adobe Research for providing the Inter-IIT Tech Meet 13.0 Problem Statement focused on detecting AI-generated images and generating interpretable explanations based on visual artifacts. The dual-task challenge, which was to identify synthetic images and provide explanations for the artifacts used in the identification, offered a valuable research opportunity. The authors also thank the organizers of Inter-IIT Tech Meet 13.0 for fostering innovation through this competition. This paper presents our independent research and solution to the tasks described above.

\bibliographystyle{plainnat}
\bibliography{references}

\begin{thebibliography}{119}
\providecommand{\natexlab}[1]{#1}
\providecommand{\url}[1]{\texttt{#1}}
\expandafter\ifx\csname urlstyle\endcsname\relax
  \providecommand{\doi}[1]{doi: #1}\else
  \providecommand{\doi}{doi: \begingroup \urlstyle{rm}\Url}\fi

\bibitem[Fak()]{FakeFormerEfficientVulnerabilityDriven}
{{FakeFormer}}: {{Efficient Vulnerability-Driven Transformers}} for {{Generalisable Deepfake Detection}}.
\newblock https://arxiv.org/html/2410.21964.

\bibitem[Abir et~al.(2022)Abir, Khanam, Alam, Hadjouni, Elmannai, Bourouis, Dey, and Khan]{abirDetectingDeepfakeImages2022}
Wahidul Abir, Faria Khanam, Kazi Alam, Myriam Hadjouni, Hela Elmannai, Sami Bourouis, Rajesh Dey, and Mohammad Khan.
\newblock Detecting {{Deepfake Images Using Deep Learning Techniques}} and {{Explainable AI Methods}}.
\newblock \emph{Intelligent Automation \& Soft Computing}, 35\penalty0 (2):\penalty0 2151--2169, 2022.
\newblock ISSN 1079-8587, 2326-005X.
\newblock \doi{10.32604/iasc.2023.029653}.

\bibitem[Agrawal et~al.(2024)Agrawal, Antoniak, Hanna, Bout, Chaplot, Chudnovsky, Costa, Monicault, Garg, Gervet, Ghosh, H{\'e}liou, Jacob, Jiang, Khandelwal, Lacroix, Lample, Casas, Lavril, Scao, Lo, Marshall, Martin, Mensch, Muddireddy, Nemychnikova, Pellat, Platen, Raghuraman, Rozi{\`e}re, Sablayrolles, Saulnier, Sauvestre, Shang, Soletskyi, Stewart, Stock, Studnia, Subramanian, Vaze, Wang, and Yang]{agrawalPixtral12B2024}
Pravesh Agrawal, Szymon Antoniak, Emma~Bou Hanna, Baptiste Bout, Devendra Chaplot, Jessica Chudnovsky, Diogo Costa, Baudouin~De Monicault, Saurabh Garg, Theophile Gervet, Soham Ghosh, Am{\'e}lie H{\'e}liou, Paul Jacob, Albert~Q. Jiang, Kartik Khandelwal, Timoth{\'e}e Lacroix, Guillaume Lample, Diego~Las Casas, Thibaut Lavril, Teven~Le Scao, Andy Lo, William Marshall, Louis Martin, Arthur Mensch, Pavankumar Muddireddy, Valera Nemychnikova, Marie Pellat, Patrick~Von Platen, Nikhil Raghuraman, Baptiste Rozi{\`e}re, Alexandre Sablayrolles, Lucile Saulnier, Romain Sauvestre, Wendy Shang, Roman Soletskyi, Lawrence Stewart, Pierre Stock, Joachim Studnia, Sandeep Subramanian, Sagar Vaze, Thomas Wang, and Sophia Yang.
\newblock Pixtral {{12B}}, October 2024.

\bibitem[Akagic et~al.(2024)Akagic, Buza, Kapo, and Bohlouli]{akagicExploringImpactReal2024}
Amila Akagic, Emir Buza, Medina Kapo, and Mahdi Bohlouli.
\newblock Exploring the {{Impact}} of {{Real}} and {{Synthetic Data}} in {{Image Classification}}: {{A Comprehensive Investigation Using CIFAKE Dataset}}.
\newblock In \emph{2024 10th {{International Conference}} on {{Control}}, {{Decision}} and {{Information Technologies}} ({{CoDIT}})}, pages 1207--1212, July 2024.
\newblock \doi{10.1109/CoDIT62066.2024.10708200}.

\bibitem[Akhtar et~al.(2021)Akhtar, Mian, Kardan, and Shah]{akhtarAdvancesAdversarialAttacks2021}
Naveed Akhtar, Ajmal Mian, Navid Kardan, and Mubarak Shah.
\newblock Advances in {{Adversarial Attacks}} and {{Defenses}} in {{Computer Vision}}: {{A Survey}}.
\newblock \emph{IEEE Access}, 9:\penalty0 155161--155196, 2021.
\newblock ISSN 2169-3536.
\newblock \doi{10.1109/ACCESS.2021.3127960}.

\bibitem[Amirian and Schwenker(2020)]{amirianRadialBasisFunction2020}
Mohammadreza Amirian and Friedhelm Schwenker.
\newblock Radial {{Basis Function Networks}} for {{Convolutional Neural Networks}} to {{Learn Similarity Distance Metric}} and {{Improve Interpretability}}.
\newblock \emph{IEEE Access}, 8:\penalty0 123087--123097, 2020.
\newblock ISSN 2169-3536.
\newblock \doi{10.1109/ACCESS.2020.3007337}.

\bibitem[Anwar and Barnes(2019)]{anwarDenselyResidualLaplacian2019}
Saeed Anwar and Nick Barnes.
\newblock Densely {{Residual Laplacian Super-Resolution}}, July 2019.

\bibitem[Ba and Caruana(2014)]{baDeepNetsReally2014}
Lei~Jimmy Ba and Rich Caruana.
\newblock Do {{Deep Nets Really Need}} to be {{Deep}}?, October 2014.

\bibitem[Bai et~al.(2025)Bai, Chen, Liu, Wang, Ge, Song, Dang, Wang, Wang, Tang, Zhong, Zhu, Yang, Li, Wan, Wang, Ding, Fu, Xu, Ye, Zhang, Xie, Cheng, Zhang, Yang, Xu, and Lin]{baiQwen25VLTechnicalReport2025}
Shuai Bai, Keqin Chen, Xuejing Liu, Jialin Wang, Wenbin Ge, Sibo Song, Kai Dang, Peng Wang, Shijie Wang, Jun Tang, Humen Zhong, Yuanzhi Zhu, Mingkun Yang, Zhaohai Li, Jianqiang Wan, Pengfei Wang, Wei Ding, Zheren Fu, Yiheng Xu, Jiabo Ye, Xi~Zhang, Tianbao Xie, Zesen Cheng, Hang Zhang, Zhibo Yang, Haiyang Xu, and Junyang Lin.
\newblock Qwen2.5-{{VL Technical Report}}, February 2025.

\bibitem[Beggel et~al.(2019)Beggel, Pfeiffer, and Bischl]{beggelRobustAnomalyDetection2019}
Laura Beggel, Michael Pfeiffer, and Bernd Bischl.
\newblock Robust {{Anomaly Detection}} in {{Images}} using {{Adversarial Autoencoders}}, January 2019.

\bibitem[Bird and Lotfi(2023)]{birdCIFAKEImageClassification2023}
Jordan~J. Bird and Ahmad Lotfi.
\newblock {{CIFAKE}}: {{Image Classification}} and {{Explainable Identification}} of {{AI-Generated Synthetic Images}}, March 2023.

\bibitem[Bituin and Antonio(2024)]{bituin_ensemble_2024}
Ronie~C Bituin and Ronielle Antonio.
\newblock Ensemble {Model} of {Lanczos} and {Bicubic} {Interpolation} with {Neural} {Network} and {Resampling} for {Image} {Enhancement}.
\newblock In \emph{Proceedings of the 2024 7th {International} {Conference} on {Software} {Engineering} and {Information} {Management}}, pages 110--115, Suva Fiji, January 2024. ACM.
\newblock \doi{10.1145/3647722.3647739}.
\newblock URL \url{https://dl.acm.org/doi/10.1145/3647722.3647739}.

\bibitem[Bommasani et~al.(2022)Bommasani, Hudson, Adeli, Altman, Arora, von Arx, Bernstein, Bohg, Bosselut, Brunskill, Brynjolfsson, Buch, Card, Castellon, Chatterji, Chen, Creel, Davis, Demszky, Donahue, Doumbouya, Durmus, Ermon, Etchemendy, Ethayarajh, {Fei-Fei}, Finn, Gale, Gillespie, Goel, Goodman, Grossman, Guha, Hashimoto, Henderson, Hewitt, Ho, Hong, Hsu, Huang, Icard, Jain, Jurafsky, Kalluri, Karamcheti, Keeling, Khani, Khattab, Koh, Krass, Krishna, Kuditipudi, Kumar, Ladhak, Lee, Lee, Leskovec, Levent, Li, Li, Ma, Malik, Manning, Mirchandani, Mitchell, Munyikwa, Nair, Narayan, Narayanan, Newman, Nie, Niebles, Nilforoshan, Nyarko, Ogut, Orr, Papadimitriou, Park, Piech, Portelance, Potts, Raghunathan, Reich, Ren, Rong, Roohani, Ruiz, Ryan, R{\'e}, Sadigh, Sagawa, Santhanam, Shih, Srinivasan, Tamkin, Taori, Thomas, Tram{\`e}r, Wang, Wang, Wu, Wu, Wu, Xie, Yasunaga, You, Zaharia, Zhang, Zhang, Zhang, Zhang, Zheng, Zhou, and Liang]{bommasaniOpportunitiesRisksFoundation2022}
Rishi Bommasani, Drew~A. Hudson, Ehsan Adeli, Russ Altman, Simran Arora, Sydney von Arx, Michael~S. Bernstein, Jeannette Bohg, Antoine Bosselut, Emma Brunskill, Erik Brynjolfsson, Shyamal Buch, Dallas Card, Rodrigo Castellon, Niladri Chatterji, Annie Chen, Kathleen Creel, Jared~Quincy Davis, Dora Demszky, Chris Donahue, Moussa Doumbouya, Esin Durmus, Stefano Ermon, John Etchemendy, Kawin Ethayarajh, Li~{Fei-Fei}, Chelsea Finn, Trevor Gale, Lauren Gillespie, Karan Goel, Noah Goodman, Shelby Grossman, Neel Guha, Tatsunori Hashimoto, Peter Henderson, John Hewitt, Daniel~E. Ho, Jenny Hong, Kyle Hsu, Jing Huang, Thomas Icard, Saahil Jain, Dan Jurafsky, Pratyusha Kalluri, Siddharth Karamcheti, Geoff Keeling, Fereshte Khani, Omar Khattab, Pang~Wei Koh, Mark Krass, Ranjay Krishna, Rohith Kuditipudi, Ananya Kumar, Faisal Ladhak, Mina Lee, Tony Lee, Jure Leskovec, Isabelle Levent, Xiang~Lisa Li, Xuechen Li, Tengyu Ma, Ali Malik, Christopher~D. Manning, Suvir Mirchandani, Eric Mitchell, Zanele Munyikwa, Suraj Nair,
  Avanika Narayan, Deepak Narayanan, Ben Newman, Allen Nie, Juan~Carlos Niebles, Hamed Nilforoshan, Julian Nyarko, Giray Ogut, Laurel Orr, Isabel Papadimitriou, Joon~Sung Park, Chris Piech, Eva Portelance, Christopher Potts, Aditi Raghunathan, Rob Reich, Hongyu Ren, Frieda Rong, Yusuf Roohani, Camilo Ruiz, Jack Ryan, Christopher R{\'e}, Dorsa Sadigh, Shiori Sagawa, Keshav Santhanam, Andy Shih, Krishnan Srinivasan, Alex Tamkin, Rohan Taori, Armin~W. Thomas, Florian Tram{\`e}r, Rose~E. Wang, William Wang, Bohan Wu, Jiajun Wu, Yuhuai Wu, Sang~Michael Xie, Michihiro Yasunaga, Jiaxuan You, Matei Zaharia, Michael Zhang, Tianyi Zhang, Xikun Zhang, Yuhui Zhang, Lucia Zheng, Kaitlyn Zhou, and Percy Liang.
\newblock On the {{Opportunities}} and {{Risks}} of {{Foundation Models}}, July 2022.

\bibitem[Brendel and Bethge(2019)]{brendelApproximatingCNNsBagoflocalFeatures2019}
Wieland Brendel and Matthias Bethge.
\newblock Approximating {{CNNs}} with {{Bag-of-local-Features}} models works surprisingly well on {{ImageNet}}, March 2019.

\bibitem[Carlini and Wagner(2017)]{carliniEvaluatingRobustnessNeural2017}
Nicholas Carlini and David Wagner.
\newblock Towards {{Evaluating}} the {{Robustness}} of {{Neural Networks}}, March 2017.

\bibitem[Chai et~al.(2020)Chai, Bau, Lim, and Isola]{chaiWhatMakesFake2020}
Lucy Chai, David Bau, Ser-Nam Lim, and Phillip Isola.
\newblock What makes fake images detectable? {{Understanding}} properties that generalize, August 2020.

\bibitem[Chang et~al.(2018)Chang, Ding, and Li]{changSingleImageSuper2018}
Kan Chang, Pak Lun~Kevin Ding, and Baoxin Li.
\newblock Single {{Image Super Resolution Using Joint Regularization}}.
\newblock \emph{IEEE Signal Processing Letters}, 25\penalty0 (4):\penalty0 596--600, April 2018.
\newblock ISSN 1558-2361.
\newblock \doi{10.1109/LSP.2018.2815003}.

\bibitem[Chen et~al.(2021)Chen, Wang, Guo, Xu, Deng, Liu, Ma, Xu, Xu, and Gao]{chenPreTrainedImageProcessing2021}
Hanting Chen, Yunhe Wang, Tianyu Guo, Chang Xu, Yiping Deng, Zhenhua Liu, Siwei Ma, Chunjing Xu, Chao Xu, and Wen Gao.
\newblock Pre-{{Trained Image Processing Transformer}}, November 2021.

\bibitem[Chen et~al.(2024)Chen, Yao, and Niu]{chenSingleSimplePatch2024}
Jiaxuan Chen, Jieteng Yao, and Li~Niu.
\newblock A {{Single Simple Patch}} is {{All You Need}} for {{AI-generated Image Detection}}, April 2024.

\bibitem[Chen et~al.(2022)Chen, Guo, Yi, Li, and Elhoseiny]{chenVisualGPTDataefficientAdaptation2022}
Jun Chen, Han Guo, Kai Yi, Boyang Li, and Mohamed Elhoseiny.
\newblock {{VisualGPT}}: {{Data-efficient Adaptation}} of {{Pretrained Language Models}} for {{Image Captioning}}.
\newblock In \emph{2022 {{IEEE}}/{{CVF Conference}} on {{Computer Vision}} and {{Pattern Recognition}} ({{CVPR}})}, pages 18009--18019, New Orleans, LA, USA, June 2022. IEEE.
\newblock ISBN 978-1-6654-6946-3.
\newblock \doi{10.1109/CVPR52688.2022.01750}.

\bibitem[Chen et~al.(2023)Chen, Zhu, Shen, Li, Liu, Zhang, Krishnamoorthi, Chandra, Xiong, and Elhoseiny]{chenMiniGPTv2LargeLanguage2023a}
Jun Chen, Deyao Zhu, Xiaoqian Shen, Xiang Li, Zechun Liu, Pengchuan Zhang, Raghuraman Krishnamoorthi, Vikas Chandra, Yunyang Xiong, and Mohamed Elhoseiny.
\newblock {{MiniGPT-v2}}: Large language model as a unified interface for vision-language multi-task learning, November 2023.

\bibitem[Coccomini et~al.(2022)Coccomini, Messina, Gennaro, and Falchi]{coccominiCombiningEfficientNetVision2022}
Davide Coccomini, Nicola Messina, Claudio Gennaro, and Fabrizio Falchi.
\newblock Combining {{EfficientNet}} and {{Vision Transformers}} for {{Video Deepfake Detection}}.
\newblock volume 13233, pages 219--229. 2022.
\newblock \doi{10.1007/978-3-031-06433-3_19}.

\bibitem[Croce and Hein(2020)]{croceReliableEvaluationAdversarial2020}
Francesco Croce and Matthias Hein.
\newblock Reliable evaluation of adversarial robustness with an ensemble of diverse parameter-free attacks, August 2020.

\bibitem[Deitke et~al.(2024)Deitke, Clark, Lee, Tripathi, Yang, Park, Salehi, Muennighoff, Lo, Soldaini, Lu, Anderson, Bransom, Ehsani, Ngo, Chen, Patel, Yatskar, {Callison-Burch}, Head, Hendrix, Bastani, VanderBilt, Lambert, Chou, Chheda, Sparks, Skjonsberg, Schmitz, Sarnat, Bischoff, Walsh, Newell, Wolters, Gupta, Zeng, Borchardt, Groeneveld, Nam, Lebrecht, Wittlif, Schoenick, Michel, Krishna, Weihs, Smith, Hajishirzi, Girshick, Farhadi, and Kembhavi]{deitkeMolmoPixMoOpen2024}
Matt Deitke, Christopher Clark, Sangho Lee, Rohun Tripathi, Yue Yang, Jae~Sung Park, Mohammadreza Salehi, Niklas Muennighoff, Kyle Lo, Luca Soldaini, Jiasen Lu, Taira Anderson, Erin Bransom, Kiana Ehsani, Huong Ngo, YenSung Chen, Ajay Patel, Mark Yatskar, Chris {Callison-Burch}, Andrew Head, Rose Hendrix, Favyen Bastani, Eli VanderBilt, Nathan Lambert, Yvonne Chou, Arnavi Chheda, Jenna Sparks, Sam Skjonsberg, Michael Schmitz, Aaron Sarnat, Byron Bischoff, Pete Walsh, Chris Newell, Piper Wolters, Tanmay Gupta, Kuo-Hao Zeng, Jon Borchardt, Dirk Groeneveld, Crystal Nam, Sophie Lebrecht, Caitlin Wittlif, Carissa Schoenick, Oscar Michel, Ranjay Krishna, Luca Weihs, Noah~A. Smith, Hannaneh Hajishirzi, Ross Girshick, Ali Farhadi, and Aniruddha Kembhavi.
\newblock Molmo and {{PixMo}}: {{Open Weights}} and {{Open Data}} for {{State-of-the-Art Vision-Language Models}}, December 2024.

\bibitem[Dhamija and Bansal(2024)]{dhamijaHowDefendSecure2024}
Lovi Dhamija and Urvashi Bansal.
\newblock How to {{Defend}} and {{Secure Deep Learning Models Against Adversarial Attacks}} in {{Computer Vision}}: {{A Systematic Review}}.
\newblock \emph{New Generation Computing}, 42\penalty0 (5):\penalty0 1165--1235, December 2024.
\newblock ISSN 0288-3635, 1882-7055.
\newblock \doi{10.1007/s00354-024-00283-0}.

\bibitem[Dong et~al.(2015)Dong, Loy, He, and Tang]{dongImageSuperResolutionUsing2015}
Chao Dong, Chen~Change Loy, Kaiming He, and Xiaoou Tang.
\newblock Image {{Super-Resolution Using Deep Convolutional Networks}}, July 2015.

\bibitem[Dong et~al.(2020)Dong, Chen, Bao, Qin, Yuan, Zhang, Yu, and Chen]{dongGreedyFoolDistortionAwareSparse2020}
Xiaoyi Dong, Dongdong Chen, Jianmin Bao, Chuan Qin, Lu~Yuan, Weiming Zhang, Nenghai Yu, and Dong Chen.
\newblock {{GreedyFool}}: {{Distortion-Aware Sparse Adversarial Attack}}, October 2020.

\bibitem[Du et~al.(2020)Du, Pentyala, Li, and Hu]{duGeneralizableDeepfakeDetection2020}
Mengnan Du, Shiva Pentyala, Yuening Li, and Xia Hu.
\newblock Towards {{Generalizable Deepfake Detection}} with {{Locality-aware AutoEncoder}}.
\newblock In \emph{Proceedings of the 29th {{ACM International Conference}} on {{Information}} \& {{Knowledge Management}}}, pages 325--334, Virtual Event Ireland, October 2020. ACM.
\newblock ISBN 978-1-4503-6859-9.
\newblock \doi{10.1145/3340531.3411892}.

\bibitem[Durall et~al.(2020)Durall, Keuper, Pfreundt, and Keuper]{durallUnmaskingDeepFakesSimple2020}
Ricard Durall, Margret Keuper, Franz-Josef Pfreundt, and Janis Keuper.
\newblock Unmasking {{DeepFakes}} with simple {{Features}}, March 2020.

\bibitem[Efros and Freeman()]{efrosImageQuiltingTexture}
Alexei~A Efros and William~T Freeman.
\newblock Image {{Quilting}} for {{Texture Synthesis}} and {{Transfer}}.

\bibitem[Emami and {Mart{\'i}nez-Mu{\~n}oz}(2023)]{emamiSequentialTrainingNeural2023}
Seyedsaman Emami and Gonzalo {Mart{\'i}nez-Mu{\~n}oz}.
\newblock Sequential {{Training}} of {{Neural Networks With Gradient Boosting}}.
\newblock \emph{IEEE Access}, 11:\penalty0 42738--42750, 2023.
\newblock ISSN 2169-3536.
\newblock \doi{10.1109/ACCESS.2023.3271515}.

\bibitem[Frank et~al.(2020)Frank, Eisenhofer, Sch{\"o}nherr, Fischer, Kolossa, and Holz]{frankLeveragingFrequencyAnalysis2020}
Joel Frank, Thorsten Eisenhofer, Lea Sch{\"o}nherr, Asja Fischer, Dorothea Kolossa, and Thorsten Holz.
\newblock Leveraging {{Frequency Analysis}} for {{Deep Fake Image Recognition}}, June 2020.

\bibitem[Geirhos et~al.(2019)Geirhos, Rubisch, Michaelis, Bethge, Wichmann, and Brendel]{geirhosImageNettrainedCNNsAre2019}
Robert Geirhos, Patricia Rubisch, Claudio Michaelis, Matthias Bethge, Felix~A. Wichmann, and Wieland Brendel.
\newblock {{ImageNet-trained CNNs}} are biased towards texture; increasing shape bias improves accuracy and robustness, January 2019.

\bibitem[Goodfellow et~al.(2014)Goodfellow, {Pouget-Abadie}, Mirza, Xu, {Warde-Farley}, Ozair, Courville, and Bengio]{goodfellowGenerativeAdversarialNetworks2014}
Ian~J. Goodfellow, Jean {Pouget-Abadie}, Mehdi Mirza, Bing Xu, David {Warde-Farley}, Sherjil Ozair, Aaron Courville, and Yoshua Bengio.
\newblock Generative {{Adversarial Networks}}, June 2014.

\bibitem[Goodfellow et~al.(2015)Goodfellow, Shlens, and Szegedy]{goodfellowExplainingHarnessingAdversarial2015a}
Ian~J. Goodfellow, Jonathon Shlens, and Christian Szegedy.
\newblock Explaining and {{Harnessing Adversarial Examples}}, March 2015.

\bibitem[Gunning et~al.(2019)Gunning, Stefik, Choi, Miller, Stumpf, and Yang]{gunning_xaiexplainable_2019}
David Gunning, Mark Stefik, Jaesik Choi, Timothy Miller, Simone Stumpf, and Guang-Zhong Yang.
\newblock {XAI}—{Explainable} artificial intelligence.
\newblock \emph{Science Robotics}, 4\penalty0 (37):\penalty0 eaay7120, 2019.
\newblock \doi{10.1126/scirobotics.aay7120}.
\newblock URL \url{https://www.science.org/doi/abs/10.1126/scirobotics.aay7120}.
\newblock \_eprint: https://www.science.org/doi/pdf/10.1126/scirobotics.aay7120.

\bibitem[Guo et~al.(2018)Guo, Rana, Cisse, and van~der Maaten]{guoCounteringAdversarialImages2018}
Chuan Guo, Mayank Rana, Moustapha Cisse, and Laurens van~der Maaten.
\newblock Countering {{Adversarial Images}} using {{Input Transformations}}, January 2018.

\bibitem[Haliassos et~al.(2021)Haliassos, Vougioukas, Petridis, and Pantic]{haliassosLipsDontLie2021}
Alexandros Haliassos, Konstantinos Vougioukas, Stavros Petridis, and Maja Pantic.
\newblock Lips {{Don}}'t {{Lie}}: {{A Generalisable}} and {{Robust Approach}} to {{Face Forgery Detection}}.
\newblock In \emph{2021 {{IEEE}}/{{CVF Conference}} on {{Computer Vision}} and {{Pattern Recognition}} ({{CVPR}})}, pages 5037--5047, Nashville, TN, USA, June 2021. IEEE.
\newblock ISBN 978-1-6654-4509-2.
\newblock \doi{10.1109/CVPR46437.2021.00500}.

\bibitem[Hannan et~al.(2025)Hannan, Cooper, White, Doster, Kvinge, and Watkins]{hannan_foundation_2025}
Darryl Hannan, John Cooper, Dylan White, Timothy Doster, Henry Kvinge, and Yijing Watkins.
\newblock Foundation {Models} for {Remote} {Sensing}: {An} {Analysis} of {MLLMs} for {Object} {Localization}, April 2025.
\newblock URL \url{http://arxiv.org/abs/2504.10727}.
\newblock arXiv:2504.10727 [cs].

\bibitem[He et~al.(2015)He, Zhang, Ren, and Sun]{heDeepResidualLearning2015}
Kaiming He, Xiangyu Zhang, Shaoqing Ren, and Jian Sun.
\newblock Deep {{Residual Learning}} for {{Image Recognition}}, December 2015.

\bibitem[Hendrycks et~al.(2023)Hendrycks, Mazeika, and Woodside]{hendrycksOverviewCatastrophicAI2023}
Dan Hendrycks, Mantas Mazeika, and Thomas Woodside.
\newblock An {{Overview}} of {{Catastrophic AI Risks}}, October 2023.

\bibitem[Hinton et~al.(2015)Hinton, Vinyals, and Dean]{hintonDistillingKnowledgeNeural2015}
Geoffrey Hinton, Oriol Vinyals, and Jeff Dean.
\newblock Distilling the {{Knowledge}} in a {{Neural Network}}, March 2015.

\bibitem[Ho et~al.(2020)Ho, Jain, and Abbeel]{hoDenoisingDiffusionProbabilistic2020}
Jonathan Ho, Ajay Jain, and Pieter Abbeel.
\newblock Denoising {{Diffusion Probabilistic Models}}, December 2020.

\bibitem[Hooda et~al.(2022)Hooda, Mangaokar, Feng, Fawaz, Jha, and Prakash]{hoodaD4DetectionAdversarial2022}
Ashish Hooda, Neal Mangaokar, Ryan Feng, Kassem Fawaz, Somesh Jha, and Atul Prakash.
\newblock D4: {{Detection}} of {{Adversarial Diffusion Deepfakes Using Disjoint Ensembles}}, 2022.

\bibitem[Hsu et~al.(2024)Hsu, Lee, and Chou]{hsuDRCTSavingImage2024}
Chih-Chung Hsu, Chia-Ming Lee, and Yi-Shiuan Chou.
\newblock {{DRCT}}: {{Saving Image Super-Resolution}} away from {{Information Bottleneck}}.
\newblock In \emph{2024 {{IEEE}}/{{CVF Conference}} on {{Computer Vision}} and {{Pattern Recognition Workshops}} ({{CVPRW}})}, pages 6133--6142, Seattle, WA, USA, June 2024. IEEE.
\newblock ISBN 979-8-3503-6547-4.
\newblock \doi{10.1109/CVPRW63382.2024.00618}.

\bibitem[Huang et~al.()Huang, Joseph, Nelson, Rubinstein, and Tygar]{huangAdversarialMachineLearning}
Ling Huang, Anthony~D Joseph, Blaine Nelson, Benjamin I~P Rubinstein, and J~D Tygar.
\newblock Adversarial {{Machine Learning}}.

\bibitem[Hwang et~al.(2019)Hwang, Park, Jang, Yoon, and Cho]{hwangPuVAEVariationalAutoencoder2019}
Uiwon Hwang, Jaewoo Park, Hyemi Jang, Sungroh Yoon, and Nam~Ik Cho.
\newblock {{PuVAE}}: {{A Variational Autoencoder}} to {{Purify Adversarial Examples}}, March 2019.

\bibitem[Jeong et~al.(2021)Jeong, Kim, Min, Joe, Gwon, and Choi]{jeongBiHPFBilateralHighPass2021}
Yonghyun Jeong, Doyeon Kim, Seungjai Min, Seongho Joe, Youngjune Gwon, and Jongwon Choi.
\newblock {{BiHPF}}: {{Bilateral High-Pass Filters}} for {{Robust Deepfake Detection}}, August 2021.

\bibitem[Jia et~al.(2022)Jia, Ma, Yao, Yin, Ding, and Yang]{jiaExploringFrequencyAdversarial2022b}
Shuai Jia, Chao Ma, Taiping Yao, Bangjie Yin, Shouhong Ding, and Xiaokang Yang.
\newblock Exploring {{Frequency Adversarial Attacks}} for {{Face Forgery Detection}}, March 2022.

\bibitem[Joe et~al.(2019)Joe, Hwang, and Shin]{joeLearningDisentangleRobust2019}
Byunggill Joe, Sung~Ju Hwang, and Insik Shin.
\newblock Learning to {{Disentangle Robust}} and {{Vulnerable Features}} for {{Adversarial Detection}}, September 2019.

\bibitem[John(2021)]{johnDiscreteCosineTransform2021}
Jacob John.
\newblock Discrete {{Cosine Transform}} in {{JPEG Compression}}, February 2021.

\bibitem[Kalaria et~al.(2022)Kalaria, Hazra, and Chakrabarti]{kalariaAdversarialPurificationUsing2022}
Dvij Kalaria, Aritra Hazra, and Partha~Pratim Chakrabarti.
\newblock Towards {{Adversarial Purification}} using {{Denoising AutoEncoders}}, August 2022.

\bibitem[Kang et~al.(2025)Kang, Wen, Wen, Ye, Li, Feng, Zhou, Wang, Lin, Zhang, and He]{kangLEGIONLearningGround2025}
Hengrui Kang, Siwei Wen, Zichen Wen, Junyan Ye, Weijia Li, Peilin Feng, Baichuan Zhou, Bin Wang, Dahua Lin, Linfeng Zhang, and Conghui He.
\newblock {{LEGION}}: {{Learning}} to {{Ground}} and {{Explain}} for {{Synthetic Image Detection}}, March 2025.

\bibitem[Keita et~al.(2024{\natexlab{a}})Keita, Hamidouche, Eutamene, Hadid, and {Taleb-Ahmed}]{keitaBiLORAVisionLanguageApproach2024}
Mamadou Keita, Wassim Hamidouche, Hessen~Bougueffa Eutamene, Abdenour Hadid, and Abdelmalik {Taleb-Ahmed}.
\newblock Bi-{{LORA}}: {{A Vision-Language Approach}} for {{Synthetic Image Detection}}, April 2024{\natexlab{a}}.

\bibitem[Keita et~al.(2024{\natexlab{b}})Keita, Hamidouche, Eutamene, {Taleb-Ahmed}, and Hadid]{keitaFIDAVLFakeImage2024}
Mamadou Keita, Wassim Hamidouche, Hessen~Bougueffa Eutamene, Abdelmalik {Taleb-Ahmed}, and Abdenour Hadid.
\newblock {{FIDAVL}}: {{Fake Image Detection}} and {{Attribution}} using {{Vision-Language Model}}, August 2024{\natexlab{b}}.

\bibitem[Khaledyan et~al.(2020)Khaledyan, Amirany, Jafari, Moaiyeri, Khuzani, and Mashhadi]{khaledyan_low-cost_2020}
Donya Khaledyan, Abdolah Amirany, Kian Jafari, Mohammad~Hossein Moaiyeri, Abolfazl~Zargari Khuzani, and Najmeh Mashhadi.
\newblock Low-{Cost} {Implementation} of {Bilinear} and {Bicubic} {Image} {Interpolation} for {Real}-{Time} {Image} {Super}-{Resolution}, September 2020.
\newblock URL \url{http://arxiv.org/abs/2009.09622}.
\newblock arXiv:2009.09622 [eess].

\bibitem[Khamaiseh et~al.(2022)Khamaiseh, Bagagem, {Al-Alaj}, Mancino, and Alomari]{khamaisehAdversarialDeepLearning2022}
Samer~Y. Khamaiseh, Derek Bagagem, Abdullah {Al-Alaj}, Mathew Mancino, and Hakam~W. Alomari.
\newblock Adversarial {{Deep Learning}}: {{A Survey}} on {{Adversarial Attacks}} and {{Defense Mechanisms}} on {{Image Classification}}.
\newblock \emph{IEEE Access}, 10:\penalty0 102266--102291, 2022.
\newblock ISSN 2169-3536.
\newblock \doi{10.1109/ACCESS.2022.3208131}.

\bibitem[Khan and {Dang-Nguyen}(2024)]{khanCLIPpingDeceptionAdapting2024}
Sohail~Ahmed Khan and Duc-Tien {Dang-Nguyen}.
\newblock {{CLIPping}} the {{Deception}}: {{Adapting Vision-Language Models}} for {{Universal Deepfake Detection}}, February 2024.

\bibitem[Khormali and Yuan(2023)]{khormaliSelfSupervisedGraphTransformer2023}
Aminollah Khormali and Jiann-Shiun Yuan.
\newblock Self-{{Supervised Graph Transformer}} for {{Deepfake Detection}}, July 2023.

\bibitem[Kingma and Welling(2022)]{kingmaAutoEncodingVariationalBayes2022}
Diederik~P. Kingma and Max Welling.
\newblock Auto-{{Encoding Variational Bayes}}, December 2022.

\bibitem[Kos et~al.(2018)Kos, Fischer, and Song]{kosAdversarialExamplesGenerative2018}
Jernej Kos, Ian Fischer, and Dawn Song.
\newblock Adversarial {{Examples}} for {{Generative Models}}.
\newblock In \emph{2018 {{IEEE Security}} and {{Privacy Workshops}} ({{SPW}})}, pages 36--42, May 2018.
\newblock \doi{10.1109/SPW.2018.00014}.

\bibitem[Kurakin et~al.(2017)Kurakin, Goodfellow, and Bengio]{kurakinAdversarialExamplesPhysical2017}
Alexey Kurakin, Ian Goodfellow, and Samy Bengio.
\newblock Adversarial examples in the physical world, February 2017.

\bibitem[Lai et~al.(2024)Lai, Yu, Yang, Li, Kang, and Shen]{laiGMDFGeneralizedMultiScenario2024}
Yingxin Lai, Zitong Yu, Jing Yang, Bin Li, Xiangui Kang, and Linlin Shen.
\newblock {{GM-DF}}: {{Generalized Multi-Scenario Deepfake Detection}}, June 2024.

\bibitem[Li et~al.(2024)Li, Wang, Li, Qian, Zhang, and Vasilakos]{liAreHandcraftedFilters2024a}
Jialiang Li, Haoyue Wang, Sheng Li, Zhenxing Qian, Xinpeng Zhang, and Athanasios~V. Vasilakos.
\newblock Are handcrafted filters helpful for attributing {{AI-generated}} images?, July 2024.

\bibitem[Li et~al.(2023)Li, Yin, Li, Chen, Wang, Ren, Li, Yang, Xu, Sun, Kong, and Liu]{liM$^3$ITLargeScaleDataset2023}
Lei Li, Yuwei Yin, Shicheng Li, Liang Chen, Peiyi Wang, Shuhuai Ren, Mukai Li, Yazheng Yang, Jingjing Xu, Xu~Sun, Lingpeng Kong, and Qi~Liu.
\newblock M\${\textasciicircum}3\${{IT}}: {{A Large-Scale Dataset}} towards {{Multi-Modal Multilingual Instruction Tuning}}, June 2023.

\bibitem[Li et~al.(2018)Li, Chang, and Lyu]{liIctuOculiExposing2018}
Yuezun Li, Ming-Ching Chang, and Siwei Lyu.
\newblock In {{Ictu Oculi}}: {{Exposing AI Created Fake Videos}} by {{Detecting Eye Blinking}}.
\newblock In \emph{2018 {{IEEE International Workshop}} on {{Information Forensics}} and {{Security}} ({{WIFS}})}, pages 1--7, Hong Kong, Hong Kong, December 2018. IEEE.
\newblock ISBN 978-1-5386-6536-7.
\newblock \doi{10.1109/WIFS.2018.8630787}.

\bibitem[Li et~al.(2025)Li, Wu, Du, Liu, Nghiem, and Shi]{liSurveyStateArt2025}
Zongxia Li, Xiyang Wu, Hongyang Du, Fuxiao Liu, Huy Nghiem, and Guangyao Shi.
\newblock A {{Survey}} of {{State}} of the {{Art Large Vision Language Models}}: {{Alignment}}, {{Benchmark}}, {{Evaluations}} and {{Challenges}}, April 2025.

\bibitem[Liang et~al.(2021)Liang, Cao, Sun, Zhang, Gool, and Timofte]{liangSwinIRImageRestoration2021}
Jingyun Liang, Jiezhang Cao, Guolei Sun, Kai Zhang, Luc~Van Gool, and Radu Timofte.
\newblock {{SwinIR}}: {{Image Restoration Using Swin Transformer}}, August 2021.

\bibitem[Liu et~al.(2023)Liu, Li, Wu, and Lee]{liuVisualInstructionTuning2023a}
Haotian Liu, Chunyuan Li, Qingyang Wu, and Yong~Jae Lee.
\newblock Visual {{Instruction Tuning}}, December 2023.

\bibitem[Liu et~al.(2024{\natexlab{a}})Liu, Li, Li, and Lee]{liuImprovedBaselinesVisual2024}
Haotian Liu, Chunyuan Li, Yuheng Li, and Yong~Jae Lee.
\newblock Improved {{Baselines}} with {{Visual Instruction Tuning}}, May 2024{\natexlab{a}}.

\bibitem[Liu et~al.(2024{\natexlab{b}})Liu, Zhang, Li, Yan, Gao, Chen, Yuan, Huang, Sun, Gao, He, and Sun]{liuSoraReviewBackground2024}
Yixin Liu, Kai Zhang, Yuan Li, Zhiling Yan, Chujie Gao, Ruoxi Chen, Zhengqing Yuan, Yue Huang, Hanchi Sun, Jianfeng Gao, Lifang He, and Lichao Sun.
\newblock Sora: {{A Review}} on {{Background}}, {{Technology}}, {{Limitations}}, and {{Opportunities}} of {{Large Vision Models}}, April 2024{\natexlab{b}}.

\bibitem[Liu et~al.(2021)Liu, Lin, Cao, Hu, Wei, Zhang, Lin, and Guo]{liuSwinTransformerHierarchical2021}
Ze~Liu, Yutong Lin, Yue Cao, Han Hu, Yixuan Wei, Zheng Zhang, Stephen Lin, and Baining Guo.
\newblock Swin {{Transformer}}: {{Hierarchical Vision Transformer}} using {{Shifted Windows}}, August 2021.

\bibitem[Lu et~al.(2023)Lu, Clark, Lee, Zhang, Khosla, Marten, Hoiem, and Kembhavi]{luUnifiedIO2Scaling2023}
Jiasen Lu, Christopher Clark, Sangho Lee, Zichen Zhang, Savya Khosla, Ryan Marten, Derek Hoiem, and Aniruddha Kembhavi.
\newblock Unified-{{IO}} 2: {{Scaling Autoregressive Multimodal Models}} with {{Vision}}, {{Language}}, {{Audio}}, and {{Action}}, December 2023.

\bibitem[Maaz et~al.(2024)Maaz, Rasheed, Khan, and Khan]{maazVideoChatGPTDetailedVideo2024}
Muhammad Maaz, Hanoona Rasheed, Salman Khan, and Fahad~Shahbaz Khan.
\newblock Video-{{ChatGPT}}: {{Towards Detailed Video Understanding}} via {{Large Vision}} and {{Language Models}}, June 2024.

\bibitem[Madry et~al.(2019)Madry, Makelov, Schmidt, Tsipras, and Vladu]{madryDeepLearningModels2019}
Aleksander Madry, Aleksandar Makelov, Ludwig Schmidt, Dimitris Tsipras, and Adrian Vladu.
\newblock Towards {{Deep Learning Models Resistant}} to {{Adversarial Attacks}}, September 2019.

\bibitem[Malolan et~al.(2020)Malolan, Parekh, and Kazi]{malolanExplainableDeepFakeDetection2020}
Badhrinarayan Malolan, Ankit Parekh, and Faruk Kazi.
\newblock Explainable {{Deep-Fake Detection Using Visual Interpretability Methods}}.
\newblock In \emph{2020 3rd {{International Conference}} on {{Information}} and {{Computer Technologies}} ({{ICICT}})}, pages 289--293, March 2020.
\newblock \doi{10.1109/ICICT50521.2020.00051}.

\bibitem[Menet et~al.(2020)Menet, Berthier, Gagnon, and M.~Fernandez]{menetSpartanNetworksSelffeaturesqueezing2020}
Fran{\c c}ois Menet, Paul Berthier, Michel Gagnon, and Jos{\'e} M.~Fernandez.
\newblock Spartan {{Networks}}: {{Self-feature-squeezing}} neural networks for increased robustness in adversarial settings.
\newblock \emph{Computers \& Security}, 88:\penalty0 101537, January 2020.
\newblock ISSN 01674048.
\newblock \doi{10.1016/j.cose.2019.05.014}.

\bibitem[Meng and Chen(2017)]{mengMagNetTwoProngedDefense2017}
Dongyu Meng and Hao Chen.
\newblock {{MagNet}}: A {{Two-Pronged Defense}} against {{Adversarial Examples}}, September 2017.

\bibitem[{Moosavi-Dezfooli} et~al.(2016){Moosavi-Dezfooli}, Fawzi, and Frossard]{moosavi-dezfooliDeepFoolSimpleAccurate2016}
Seyed-Mohsen {Moosavi-Dezfooli}, Alhussein Fawzi, and Pascal Frossard.
\newblock {{DeepFool}}: {{A Simple}} and {{Accurate Method}} to {{Fool Deep Neural Networks}}.
\newblock In \emph{2016 {{IEEE Conference}} on {{Computer Vision}} and {{Pattern Recognition}} ({{CVPR}})}, pages 2574--2582, June 2016.
\newblock \doi{10.1109/CVPR.2016.282}.

\bibitem[Naskar et~al.(2024)Naskar, Mohiuddin, Malakar, Cuevas, and Sarkar]{naskarDeepfakeDetectionUsing2024}
Gourab Naskar, Sk~Mohiuddin, Samir Malakar, Erik Cuevas, and Ram Sarkar.
\newblock Deepfake detection using deep feature stacking and meta-learning.
\newblock \emph{Heliyon}, 10\penalty0 (4):\penalty0 e25933, February 2024.
\newblock ISSN 24058440.
\newblock \doi{10.1016/j.heliyon.2024.e25933}.

\bibitem[Ojha et~al.(2024)Ojha, Li, and Lee]{ojhaUniversalFakeImage2024}
Utkarsh Ojha, Yuheng Li, and Yong~Jae Lee.
\newblock Towards {{Universal Fake Image Detectors}} that {{Generalize Across Generative Models}}, April 2024.

\bibitem[Papernot et~al.(2016{\natexlab{a}})Papernot, McDaniel, Jha, Fredrikson, Celik, and Swami]{papernotLimitationsDeepLearning2016}
Nicolas Papernot, Patrick McDaniel, Somesh Jha, Matt Fredrikson, Z.~Berkay Celik, and Ananthram Swami.
\newblock The {{Limitations}} of {{Deep Learning}} in {{Adversarial Settings}}.
\newblock In \emph{2016 {{IEEE European Symposium}} on {{Security}} and {{Privacy}} ({{EuroS}}\&{{P}})}, pages 372--387, March 2016{\natexlab{a}}.
\newblock \doi{10.1109/EuroSP.2016.36}.

\bibitem[Papernot et~al.(2016{\natexlab{b}})Papernot, McDaniel, Wu, Jha, and Swami]{papernotDistillationDefenseAdversarial2016}
Nicolas Papernot, Patrick McDaniel, Xi~Wu, Somesh Jha, and Ananthram Swami.
\newblock Distillation as a {{Defense}} to {{Adversarial Perturbations Against Deep Neural Networks}}.
\newblock In \emph{2016 {{IEEE Symposium}} on {{Security}} and {{Privacy}} ({{SP}})}, pages 582--597, May 2016{\natexlab{b}}.
\newblock \doi{10.1109/SP.2016.41}.

\bibitem[Pinhasov et~al.(2024)Pinhasov, Lapid, Ohayon, Sipper, and Aperstein]{pinhasovXAIBasedDetectionAdversarial2024}
Ben Pinhasov, Raz Lapid, Rony Ohayon, Moshe Sipper, and Yehudit Aperstein.
\newblock {{XAI-Based Detection}} of {{Adversarial Attacks}} on {{Deepfake Detectors}}, August 2024.

\bibitem[Qian et~al.(2020)Qian, Yin, Sheng, Chen, and Shao]{qianThinkingFrequencyFace2020}
Yuyang Qian, Guojun Yin, Lu~Sheng, Zixuan Chen, and Jing Shao.
\newblock Thinking in {{Frequency}}: {{Face Forgery Detection}} by {{Mining Frequency-Aware Clues}}.
\newblock In Andrea Vedaldi, Horst Bischof, Thomas Brox, and Jan-Michael Frahm, editors, \emph{Computer {{Vision}} -- {{ECCV}} 2020}, volume 12357, pages 86--103. Springer International Publishing, Cham, 2020.
\newblock ISBN 978-3-030-58609-6 978-3-030-58610-2.
\newblock \doi{10.1007/978-3-030-58610-2_6}.

\bibitem[Qin et~al.(2019)Qin, Martens, Gowal, Krishnan, Dvijotham, Fawzi, De, Stanforth, and Kohli]{qinAdversarialRobustnessLocal2019}
Chongli Qin, James Martens, Sven Gowal, Dilip Krishnan, Krishnamurthy Dvijotham, Alhussein Fawzi, Soham De, Robert Stanforth, and Pushmeet Kohli.
\newblock Adversarial {{Robustness}} through {{Local Linearization}}, October 2019.

\bibitem[Qin et~al.(2023)Qin, Ji, Khan, Fan, Khan, and Gool]{qinHowGoodGoogle2023}
Haotong Qin, Ge-Peng Ji, Salman Khan, Deng-Ping Fan, Fahad~Shahbaz Khan, and Luc~Van Gool.
\newblock How {{Good}} is {{Google Bard}}'s {{Visual Understanding}}? {{An Empirical Study}} on {{Open Challenges}}.
\newblock \emph{Machine Intelligence Research}, 20\penalty0 (5):\penalty0 605--613, October 2023.
\newblock ISSN 2731-538X, 2731-5398.
\newblock \doi{10.1007/s11633-023-1469-x}.

\bibitem[Radford et~al.(2021)Radford, Kim, Hallacy, Ramesh, Goh, Agarwal, Sastry, Askell, Mishkin, Clark, Krueger, and Sutskever]{radfordLearningTransferableVisual2021a}
Alec Radford, Jong~Wook Kim, Chris Hallacy, Aditya Ramesh, Gabriel Goh, Sandhini Agarwal, Girish Sastry, Amanda Askell, Pamela Mishkin, Jack Clark, Gretchen Krueger, and Ilya Sutskever.
\newblock Learning {{Transferable Visual Models From Natural Language Supervision}}, February 2021.

\bibitem[Rombach et~al.(2022)Rombach, Blattmann, Lorenz, Esser, and Ommer]{rombachHighResolutionImageSynthesis2022}
Robin Rombach, Andreas Blattmann, Dominik Lorenz, Patrick Esser, and Bj{\"o}rn Ommer.
\newblock High-{{Resolution Image Synthesis}} with {{Latent Diffusion Models}}, April 2022.

\bibitem[Romeo et~al.(2024)Romeo, Federico, Anxhelo, Raoul, and Luigi]{romeoFasterLiesRealtime2024}
Lanzino Romeo, Fontana Federico, Diko Anxhelo, Marini~Marco Raoul, and Cinque Luigi.
\newblock Faster {{Than Lies}}: {{Real-time Deepfake Detection}} using {{Binary Neural Networks}}, June 2024.

\bibitem[Ross and {Doshi-Velez}(2018)]{rossImprovingAdversarialRobustness2018}
Andrew Ross and Finale {Doshi-Velez}.
\newblock Improving the {{Adversarial Robustness}} and {{Interpretability}} of {{Deep Neural Networks}} by {{Regularizing Their Input Gradients}}.
\newblock \emph{Proceedings of the AAAI Conference on Artificial Intelligence}, 32\penalty0 (1), April 2018.
\newblock ISSN 2374-3468, 2159-5399.
\newblock \doi{10.1609/aaai.v32i1.11504}.

\bibitem[R{\"o}ssler et~al.(2019)R{\"o}ssler, Cozzolino, Verdoliva, Riess, Thies, and Nie{\ss}ner]{rosslerFaceForensicsLearningDetect2019}
Andreas R{\"o}ssler, Davide Cozzolino, Luisa Verdoliva, Christian Riess, Justus Thies, and Matthias Nie{\ss}ner.
\newblock {{FaceForensics}}++: {{Learning}} to {{Detect Manipulated Facial Images}}, August 2019.

\bibitem[Rudin et~al.(1992)Rudin, Osher, and Fatemi]{rudinNonlinearTotalVariation1992}
Leonid~I. Rudin, Stanley Osher, and Emad Fatemi.
\newblock Nonlinear total variation based noise removal algorithms.
\newblock \emph{Physica D: Nonlinear Phenomena}, 60\penalty0 (1-4):\penalty0 259--268, November 1992.
\newblock ISSN 01672789.
\newblock \doi{10.1016/0167-2789(92)90242-F}.

\bibitem[Selvaraju et~al.(2020)Selvaraju, Cogswell, Das, Vedantam, Parikh, and Batra]{selvarajuGradCAMVisualExplanations2020}
Ramprasaath~R. Selvaraju, Michael Cogswell, Abhishek Das, Ramakrishna Vedantam, Devi Parikh, and Dhruv Batra.
\newblock Grad-{{CAM}}: {{Visual Explanations}} from {{Deep Networks}} via {{Gradient-based Localization}}.
\newblock \emph{International Journal of Computer Vision}, 128\penalty0 (2):\penalty0 336--359, February 2020.
\newblock ISSN 0920-5691, 1573-1405.
\newblock \doi{10.1007/s11263-019-01228-7}.

\bibitem[Sun et~al.(2019)Sun, Tsai, Liu, Yu, and Su]{sunAdversarialDefenseStratified2019}
Bo~Sun, Nian-Hsuan Tsai, Fangchen Liu, Ronald Yu, and Hao Su.
\newblock Adversarial {{Defense}} by {{Stratified Convolutional Sparse Coding}}.
\newblock In \emph{2019 {{IEEE}}/{{CVF Conference}} on {{Computer Vision}} and {{Pattern Recognition}} ({{CVPR}})}, pages 11439--11448, Long Beach, CA, USA, June 2019. IEEE.
\newblock ISBN 978-1-7281-3293-8.
\newblock \doi{10.1109/CVPR.2019.01171}.

\bibitem[Sun and Li(2019)]{sunSuperResolutionReconstruction2019}
Na~Sun and Huina Li.
\newblock Super {{Resolution Reconstruction}} of {{Images Based}} on {{Interpolation}} and {{Full Convolutional Neural Network}} and {{Application}} in {{Medical Fields}}.
\newblock \emph{IEEE Access}, 7:\penalty0 186470--186479, 2019.
\newblock ISSN 2169-3536.
\newblock \doi{10.1109/ACCESS.2019.2960828}.

\bibitem[Szegedy et~al.(2014)Szegedy, Zaremba, Sutskever, Bruna, Erhan, Goodfellow, and Fergus]{szegedyIntriguingPropertiesNeural2014}
Christian Szegedy, Wojciech Zaremba, Ilya Sutskever, Joan Bruna, Dumitru Erhan, Ian Goodfellow, and Rob Fergus.
\newblock Intriguing properties of neural networks, February 2014.

\bibitem[Tan et~al.(2023)Tan, Zhao, Wei, Gu, and Wei]{tanLearningGradientsGeneralized2023}
Chuangchuang Tan, Yao Zhao, Shikui Wei, Guanghua Gu, and Yunchao Wei.
\newblock Learning on {{Gradients}}: {{Generalized Artifacts Representation}} for {{GAN-Generated Images Detection}}.
\newblock In \emph{2023 {{IEEE}}/{{CVF Conference}} on {{Computer Vision}} and {{Pattern Recognition}} ({{CVPR}})}, pages 12105--12114, Vancouver, BC, Canada, June 2023. IEEE.
\newblock ISBN 979-8-3503-0129-8.
\newblock \doi{10.1109/CVPR52729.2023.01165}.

\bibitem[Tishby and Zaslavsky(2015)]{tishbyDeepLearningInformation2015}
Naftali Tishby and Noga Zaslavsky.
\newblock Deep learning and the information bottleneck principle.
\newblock In \emph{2015 {{IEEE Information Theory Workshop}} ({{ITW}})}, pages 1--5, Jerusalem, Israel, April 2015. IEEE.
\newblock ISBN 978-1-4799-5524-4 978-1-4799-5526-8.
\newblock \doi{10.1109/ITW.2015.7133169}.

\bibitem[Tong and Leung(2007)]{tongSuperresolutionReconstructionBased2007}
C.~S. Tong and K.~T. Leung.
\newblock Super-resolution reconstruction based on linear interpolation of wavelet coefficients.
\newblock \emph{Multidimensional Systems and Signal Processing}, 18\penalty0 (2-3):\penalty0 153--171, September 2007.
\newblock ISSN 0923-6082, 1573-0824.
\newblock \doi{10.1007/s11045-007-0023-2}.

\bibitem[{van der Maaten} and Hinton(2008)]{JMLR:v9:vandermaaten08a}
Laurens {van der Maaten} and Geoffrey Hinton.
\newblock Visualizing data using t-{{SNE}}.
\newblock \emph{Journal of Machine Learning Research}, 9\penalty0 (86):\penalty0 2579--2605, 2008.

\bibitem[Wang et~al.(2021)Wang, Cun, Bao, Zhou, Liu, and Li]{wangUformerGeneralUShaped2021}
Zhendong Wang, Xiaodong Cun, Jianmin Bao, Wengang Zhou, Jianzhuang Liu, and Houqiang Li.
\newblock Uformer: {{A General U-Shaped Transformer}} for {{Image Restoration}}, November 2021.

\bibitem[Wang et~al.(2024)Wang, Cheng, Xiong, Xu, Li, Veeravalli, and Yang]{wangTimelySurveyVision2024}
Zhikan Wang, Zhongyao Cheng, Jiajie Xiong, Xun Xu, Tianrui Li, Bharadwaj Veeravalli, and Xulei Yang.
\newblock A {{Timely Survey}} on {{Vision Transformer}} for {{Deepfake Detection}}, May 2024.

\bibitem[Wu et~al.(2025)Wu, Zhang, Shi, Yin, Wang, Gan, Wang, Lv, Zheng, and Huang]{wuExplainableSyntheticImage2025}
Yixin Wu, Feiran Zhang, Tianyuan Shi, Ruicheng Yin, Zhenghua Wang, Zhenliang Gan, Xiaohua Wang, Changze Lv, Xiaoqing Zheng, and Xuanjing Huang.
\newblock Explainable {{Synthetic Image Detection}} through {{Diffusion Timestep Ensembling}}, March 2025.

\bibitem[Xiang et~al.(2021)Xiang, Bhagoji, Sehwag, and Mittal]{xiangPatchGuardProvablyRobust2021}
Chong Xiang, Arjun~Nitin Bhagoji, Vikash Sehwag, and Prateek Mittal.
\newblock {{PatchGuard}}: {{A Provably Robust Defense}} against {{Adversarial Patches}} via {{Small Receptive Fields}} and {{Masking}}, March 2021.

\bibitem[Xu et~al.(2020)Xu, Shi, Zhang, Wang, Chang, Huang, Kailkhura, Lin, and Hsieh]{xuAutomaticPerturbationAnalysis2020}
Kaidi Xu, Zhouxing Shi, Huan Zhang, Yihan Wang, Kai-Wei Chang, Minlie Huang, Bhavya Kailkhura, Xue Lin, and Cho-Jui Hsieh.
\newblock Automatic {{Perturbation Analysis}} for {{Scalable Certified Robustness}} and {{Beyond}}, October 2020.

\bibitem[Yan et~al.(2021)Yan, Li, Dai, Bai, and Xia]{yanD2DefendDualDomainBased2021}
Xin Yan, Yanjie Li, Tao Dai, Yang Bai, and Shu-Tao Xia.
\newblock {{D2Defend}}: {{Dual-Domain}} based {{Defense}} against {{Adversarial Examples}}.
\newblock In \emph{2021 {{International Joint Conference}} on {{Neural Networks}} ({{IJCNN}})}, pages 1--8, July 2021.
\newblock \doi{10.1109/IJCNN52387.2021.9533589}.

\bibitem[Yin et~al.(2024)Yin, Fu, Zhao, Xu, Wang, Sui, Shen, Li, Sun, and Chen]{yinWoodpeckerHallucinationCorrection2024}
Shukang Yin, Chaoyou Fu, Sirui Zhao, Tong Xu, Hao Wang, Dianbo Sui, Yunhang Shen, Ke~Li, Xing Sun, and Enhong Chen.
\newblock Woodpecker: {{Hallucination Correction}} for {{Multimodal Large Language Models}}.
\newblock \emph{Science China Information Sciences}, 67\penalty0 (12):\penalty0 220105, December 2024.
\newblock ISSN 1674-733X, 1869-1919.
\newblock \doi{10.1007/s11432-024-4251-x}.

\bibitem[Yu et~al.(2017)Yu, Cao, He, Sun, and Dai]{yuSingleimageSuperresolutionBased2017}
Lejun Yu, Siming Cao, Jun He, Bo~Sun, and Feng Dai.
\newblock Single-image super-resolution based on regularization with stationary gradient fidelity.
\newblock In \emph{2017 10th {{International Congress}} on {{Image}} and {{Signal Processing}}, {{BioMedical Engineering}} and {{Informatics}} ({{CISP-BMEI}})}, pages 1--5, October 2017.
\newblock \doi{10.1109/CISP-BMEI.2017.8301942}.

\bibitem[Yu et~al.(2019)Yu, Davis, and Fritz]{yuAttributingFakeImages2019}
Ning Yu, Larry Davis, and Mario Fritz.
\newblock Attributing {{Fake Images}} to {{GANs}}: {{Learning}} and {{Analyzing GAN Fingerprints}}.
\newblock In \emph{2019 {{IEEE}}/{{CVF International Conference}} on {{Computer Vision}} ({{ICCV}})}, pages 7555--7565, Seoul, Korea (South), October 2019. IEEE.
\newblock ISBN 978-1-7281-4803-8.
\newblock \doi{10.1109/ICCV.2019.00765}.

\bibitem[Zhang et~al.(2024{\natexlab{a}})Zhang, Yu, Dong, Li, Su, Chu, and Yu]{zhangMMLLMsRecentAdvances2024}
Duzhen Zhang, Yahan Yu, Jiahua Dong, Chenxing Li, Dan Su, Chenhui Chu, and Dong Yu.
\newblock {{MM-LLMs}}: {{Recent Advances}} in {{MultiModal Large Language Models}}.
\newblock In Lun-Wei Ku, Andre Martins, and Vivek Srikumar, editors, \emph{Findings of the {{Association}} for {{Computational Linguistics}}: {{ACL}} 2024}, pages 12401--12430, Bangkok, Thailand, August 2024{\natexlab{a}}. Association for Computational Linguistics.
\newblock \doi{10.18653/v1/2024.findings-acl.738}.

\bibitem[Zhang et~al.(2024{\natexlab{b}})Zhang, Zhao, Lim, Asadi, Huang, Yu, and Gao]{zhangVideoDeepfakeClassification2024}
Li~Zhang, Dezong Zhao, Chee~Peng Lim, Houshyar Asadi, Haoqian Huang, Yonghong Yu, and Rong Gao.
\newblock Video {{Deepfake}} classification using particle swarm optimization-based evolving ensemble models.
\newblock \emph{Knowledge-Based Systems}, 289:\penalty0 111461, April 2024{\natexlab{b}}.
\newblock ISSN 09507051.
\newblock \doi{10.1016/j.knosys.2024.111461}.

\bibitem[Zhang et~al.(2018)Zhang, Fan, Bao, Liu, and Zhang]{zhangSingleImageSuperResolutionBased2018}
Yunfeng Zhang, Qinglan Fan, Fangxun Bao, Yifang Liu, and Caiming Zhang.
\newblock Single-{{Image Super-Resolution Based}} on {{Rational Fractal Interpolation}}.
\newblock \emph{IEEE Transactions on Image Processing}, 27\penalty0 (8):\penalty0 3782--3797, August 2018.
\newblock ISSN 1941-0042.
\newblock \doi{10.1109/TIP.2018.2826139}.

\bibitem[Zhang et~al.(2023)Zhang, Li, Li, and Guo]{zhangReviewAdversarialAttacks2023}
Yutong Zhang, Yao Li, Yin Li, and Zhichang Guo.
\newblock A {{Review}} of {{Adversarial Attacks}} in {{Computer Vision}}, August 2023.

\bibitem[Zhang et~al.(2020)Zhang, Yuan, McCoyd, and Wagner]{zhangClippedBagNetDefending2020}
Zhanyuan Zhang, Benson Yuan, Michael McCoyd, and David Wagner.
\newblock Clipped {{BagNet}}: {{Defending Against Sticker Attacks}} with {{Clipped Bag-of-features}}.
\newblock In \emph{2020 {{IEEE Security}} and {{Privacy Workshops}} ({{SPW}})}, pages 55--61, May 2020.
\newblock \doi{10.1109/SPW50608.2020.00026}.

\bibitem[Zhang et~al.(2024{\natexlab{c}})Zhang, Zeng, Liu, and Zhou]{zhangNovelPerspectiveAdversarial2024}
Zhun Zhang, Yi~Zeng, Qihe Liu, and Shijie Zhou.
\newblock Towards a {{Novel Perspective}} on {{Adversarial Examples Driven}} by {{Frequency}}, April 2024{\natexlab{c}}.

\bibitem[Zhong et~al.(2024)Zhong, Xu, Li, Qian, and Zhang]{zhongPatchCraftExploringTexture2024}
Nan Zhong, Yiran Xu, Sheng Li, Zhenxing Qian, and Xinpeng Zhang.
\newblock {{PatchCraft}}: {{Exploring Texture Patch}} for {{Efficient AI-generated Image Detection}}, March 2024.

\bibitem[Zhong and Deng(2019)]{zhongAdversarialLearningMarginBased2019}
Yaoyao Zhong and Weihong Deng.
\newblock Adversarial {{Learning With Margin-Based Triplet Embedding Regularization}}.
\newblock In \emph{2019 {{IEEE}}/{{CVF International Conference}} on {{Computer Vision}} ({{ICCV}})}, pages 6548--6557, Seoul, Korea (South), October 2019. IEEE.
\newblock ISBN 978-1-7281-4803-8.
\newblock \doi{10.1109/ICCV.2019.00665}.

\bibitem[Zhu et~al.()Zhu, Chen, Shen, Li, and Elhoseiny]{zhuMINIGPT4ENHANCINGVISIONLANGUAGE}
Deyao Zhu, Jun Chen, Xiaoqian Shen, Xiang Li, and Mohamed Elhoseiny.
\newblock {{MINIGPT-4}}: {{ENHANCING VISION-LANGUAGE UNDERSTANDING WITH ADVANCED LARGE LANGUAGE MODELS}}.

\end{thebibliography}
%
%




%
\clearpage
\appendix
\renewcommand{\thesection}{Appendix \Alph{section}}
\section{Artifact Prompting}
\textbf{Instruction:} You are a helpful assistant that identifies errors and artifacts in images. Given the error code, describe instances in the image where the error occurs.

\textbf{JSON Schema:}
\begin{quote}
\ttfamily
\{“artifact”: “...”, “description”: “...” \}
\end{quote}

\textbf{Example:}
\begin{quote}
\ttfamily
\{“artifact”: “biological\_asymmetry”, “description”: “In the given image, the horse has unsymmetrical eyes” \}
\end{quote}

\textbf{Guidelines:}
\begin{itemize}
    \item Only describe the given artifact. Do not mention unrelated defects.
    \item Limit each response to 1–2 lines.
    \item Use directional or anatomical terms (e.g., ``left paw,'' ``lower trunk'').
    \item Highlight visibility using terms like ``noticeable,'' ``clearly seen,'' or ``subtle.''
    \item Follow the JSON schema strictly.
\end{itemize}

\section{Adversarial Attacks and Defense Mechanisms}
Adversarial attacks on images involve applying carefully crafted perturbations to mislead image classification models. While some attacks are carried out without significant knowledge of the weights, gradients and other features of the target model, such as generic modifications to the spatial or frequency-domain features of the image (black-box attacks), other attacks are designed to affect the classification outcome of a particular model based upon limited or perfect knowledge of the workings of the target model(gray-box or white-box attacks, respectively) \citep{khamaisehAdversarialDeepLearning2022} \citep{zhangReviewAdversarialAttacks2023}, \citep{dhamijaHowDefendSecure2024}. Such attacks have been shown to exploit various vulnerabilities in models and pose serious security risks in practical settings \citep{goodfellowExplainingHarnessingAdversarial2015a}, \citep{huangAdversarialMachineLearning}, \citep{szegedyIntriguingPropertiesNeural2014}, \citep{papernotLimitationsDeepLearning2016}, \citep{kurakinAdversarialExamplesPhysical2017}. Such attacks help enhance machine learning models by exposing vulnerabilities and, if incorporated in the training process, encouraging reliance on meaningful features instead of secondary details.\\

Certain white-box attacks exploit the gradients of the target model, such as the Fast Gradient Sign Method (FGSM) \citep{goodfellowExplainingHarnessingAdversarial2015a}, DeepFool (DF) \citep{moosavi-dezfooliDeepFoolSimpleAccurate2016} and Jacobian-based Saliency Map Attacks (JSMA) \citep{papernotLimitationsDeepLearning2016}, with more advanced attacks also being proposed such as GreedyFool \citep{dongGreedyFoolDistortionAwareSparse2020}. Other white-box attacks convert attacks into optimization problems, such as the Carlini and Wagner Attack (C\&W) \citep{carliniEvaluatingRobustnessNeural2017}. Grey-box attacks include the usage of generative models such as Variational AutoEncoders to generate targeted adversarial examples \citep{kosAdversarialExamplesGenerative2018}. Black-box attacks are usually executed as query-based attacks or transfer-based attacks, which operate solely based on outputs of the model and no knowledge of its inner workings \citep{akhtarAdvancesAdversarialAttacks2021} \citep{dhamijaHowDefendSecure2024}. Some attacks are discussed in greater detail below:
\paragraph{Fast Gradient Sign Method} \citep{goodfellowExplainingHarnessingAdversarial2015a}: It exploits the gradient of the loss function with respect to the input data to craft a perturbation in the direction that maximizes the model’s loss.

\begin{algorithm}[H]
\caption{Fast Gradient Sign Method (FGSM)}
\label{alg:fgsm}
\begin{algorithmic}
\STATE \textbf{Input}: Input $x$, true label $y$, model parameters $\theta$, loss function $J(\theta, x, y)$, perturbation magnitude $\epsilon$
\STATE \textbf{Output}: Adversarial example $x'$
\STATE Compute gradient: $\mathbf{g} \leftarrow \nabla_x J(\theta, x, y)$
\STATE Compute perturbation: $\eta \leftarrow \epsilon \cdot \text{sign}(\mathbf{g})$
\STATE Generate adversarial example: $x' \leftarrow x + \eta$
\STATE \textbf{return} $x'$
\end{algorithmic}
\end{algorithm}

\paragraph{Projected Gradient Descent}  \citep{madryDeepLearningModels2019}: It is an iterative adversarial attack that builds upon FGSM by applying multiple small gradient steps, followed by a projection onto the allowed perturbation space to ensure the perturbation remains within a defined ball around the original input. 
\begin{algorithm}[H]
\caption{Projected Gradient Descent (PGD) Attack}
\label{alg:pgd}
\textbf{Input}: Input $x$, true label $y$, loss function $J(\theta, x, y)$, perturbation magnitude $\epsilon$, step size $\alpha$, iterations $T$ \\
\textbf{Output}: Adversarial example $x'$
\begin{algorithmic}[1]
\STATE Initialize $x_0 \leftarrow x$
\FOR{$t = 1$ to $T$}
    \STATE Compute gradient: $\mathbf{g}_t \leftarrow \nabla_x J(\theta, x_t, y)$
    \STATE Update input: $x_{t+1} \leftarrow x_t + \alpha \cdot \text{sign}(\mathbf{g}_t)$
    \STATE Project: $x_{t+1} \leftarrow \text{Clip}(x_{t+1}, x - \epsilon, x + \epsilon)$
\ENDFOR
\STATE \textbf{return} $x' \leftarrow x_T$
\end{algorithmic}[1]
\end{algorithm}

\paragraph{Wavelet Packet Decomposition} \citep{zhangNovelPerspectiveAdversarial2024}: This method uses a wavelet transform to break down an input image into frequency components. By slightly altering the wavelet coefficients at specific scales and then reconstructing the image, it creates an adversarial example designed to mislead neural networks.
\begin{algorithm}[H]
\caption{Wavelet Decomposition Attack}
\label{alg:wavelet}
\textbf{Input}: Input image $x$, true label $y$, loss function $J(\theta, x, y)$, perturbation magnitude $\epsilon$, wavelet levels $L$ \\
\textbf{Output}: Adversarial example $x'$
\begin{algorithmic}[1]
\STATE Compute wavelet decomposition: $W \leftarrow \text{WaveletDecompose}(x)$
\STATE Compute gradient: $\mathbf{g}_W \leftarrow \nabla_W J(\theta, W, y)$
\STATE Select coefficients from first $L$ levels
\STATE Perturb selected coefficients: $W' \leftarrow W + \epsilon \cdot \mathbf{g}_W$
\STATE Reconstruct image: $x' \leftarrow \text{InverseWavelet}(W')$
\STATE \textbf{return} $x'$
\end{algorithmic}
\end{algorithm}

\paragraph{AutoAttack} \citep{croceReliableEvaluationAdversarial2020}: AutoAttack is an ensemble-based method designed for untargeted adversarial attacks, aiming to generate adversarial examples that cause a model to misclassify without targeting a specific class. By adaptively combining multiple attack strategies, AutoAttack achieves a high success rate of confusing models with lesser magnitude of applied perturbations. 

\begin{algorithm}[H]
\caption{AutoAttack}
\label{alg:autoattack}
\textbf{Input}: Input image $x$, true label $y$, model $f$, perturbation magnitude $\epsilon$ \\
\textbf{Output}: Adversarial example $x'$
\begin{algorithmic}[1]
\STATE Initialize $x_0 \leftarrow x$
\STATE Select predefined set of attack methods
\FOR{each attack method $\mathcal{A}$ in the set}
    \STATE Generate attack: $x' \leftarrow \mathcal{A}(x_0, y, f, \epsilon)$
    \IF{$x'$ is adversarial}
        \STATE \textbf{return} $x'$
    \ENDIF
\ENDFOR
\STATE \textbf{return} $x'$ (final adversarial example)
\end{algorithmic}
\end{algorithm}

Traditional spatial domain attacks, such as FGSM and PGD, often result in noticeable artifacts that can be easily detected by both humans and detectors. In contrast, frequency-domain attacks, particularly those manipulating high-frequency bands using Discrete Cosine Transform (DCT), allow for imperceptible perturbations that usually leave the image intact visually, making them difficult to identify. A hybrid attack strategy combining both spatial and frequency domains has shown commendable promise in the disruption of model classification \citep{jiaExploringFrequencyAdversarial2022b}.\\

Improving robustness to adversarial examples has been an active area of research. Some significant directions of improving robustness are adversarial training, image transformations, and image denoising \citep{goodfellowExplainingHarnessingAdversarial2015a}, \citep{szegedyIntriguingPropertiesNeural2014}, \citep{madryDeepLearningModels2019}, \citep{guoCounteringAdversarialImages2018}. Adversarial training involves creating a training dataset that contains a variety of adversarial attacks and enabling the model to recognize and accurately classify such perturbations \citep{goodfellowExplainingHarnessingAdversarial2015a}, \citep{szegedyIntriguingPropertiesNeural2014}, \citep{madryDeepLearningModels2019},  \citep{kurakinAdversarialExamplesPhysical2017}. To tackle the issue of computational costs involved with adversarial training of large models and a wide range of attacks, \citep{qinAdversarialRobustnessLocal2019} proposed a method that simulates adversarial training on the loss surface using a new regularizer. Another method of input transformation involves Feature Representation learning to differentiate between different features better. \citep{joeLearningDisentangleRobust2019} encourages the model to learn the feature space to identify vulnerable latent features, which form the basis for the effectiveness of adversarial attacks.\\

\citep{menetSpartanNetworksSelffeaturesqueezing2020} proposed Spartan Networks, which are deep neural networks with a modified activation function that reduces the information obtained from the input, hence forcing only relevant features to be learned. Since reducing the amount of excess information has a positive impact on robustness, BAGNet \citep{brendelApproximatingCNNsBagoflocalFeatures2019} and its derivatives, such as Clipped BAGNet \citep{zhangClippedBagNetDefending2020}, limit the receptive field of the convolutional layers through dimension reduction of filters and introducing averaging methods instead. PatchGuard \citep{xiangPatchGuardProvablyRobust2021} also introduces a robust masking scheme to mask attacks on images, apart from reducing the input receptive field. \citep{goodfellowExplainingHarnessingAdversarial2015a}, \citep{szegedyIntriguingPropertiesNeural2014}, \citep{madryDeepLearningModels2019} proposed stochastic input transformations such as bit-depth reduction and JPEG compression, which can be applied to adversarially attacked images before feeding them into convolutional networks, and demonstrated that total variance minimization \citep{rudinNonlinearTotalVariation1992} and image quilting \citep{efrosImageQuiltingTexture} are effective in tackling many adversarial attacks. Defensive distillation, proposed by \citep{papernotDistillationDefenseAdversarial2016} and built upon the idea of distillation ideated by \citep{baDeepNetsReally2014}, uses an innovative method of training a “teacher” model on the input data, following which the output probabilities are learned by a “student” model, which helps the student model make smoother decisions and evade attacks. Transitioning from linear models to nonlinear models such as Radial Basis Function (RBF) Networks \citep{amirianRadialBasisFunction2020} has shown improved resilience against adversarial examples due to their local receptive properties. Since several attacks utilize gradient information from models, and since model sensitivity is related to how small changes in the input can affect the outcome of the classification, \citep{rossImprovingAdversarialRobustness2018} performs input regularization, which then reduces the tendency of target models to flip decisions with perturbed inputs. Other forms of regularization are also explored in \citep{zhongAdversarialLearningMarginBased2019}, \citep{qinAdversarialRobustnessLocal2019}.\\ 

To target imperceptible manipulations in the frequency domain, some approaches decompose images into high and low-frequency components, highlighting forgery-prone regions such as edges and textures. This decomposition, along with methods such as Fourier analysis, also helps identify frequency-domain artifacts and signatures of generative models, such as GANs \citep{durallUnmaskingDeepFakesSimple2020}, \citep{frankLeveragingFrequencyAnalysis2020}. \citep{qianThinkingFrequencyFace2020} propose F3-Net, combining Frequency-Aware Decomposition (FAD) and Local Frequency Statistics (LFS). FAD uses bandpass filtering to isolate high-frequency components, where manipulations like edges or textures are more evident, while LFS applies a sliding-window Discrete Cosine Transform (DCT) \citep{johnDiscreteCosineTransform2021} to detect texture inconsistencies. An alternative approach to adversarial robustness is ensuring that the internal representations and feature maps of the model are modified to be resilient to perturbations. This is done by \citep{sunAdversarialDefenseStratified2019}, which introduces a sparse transformation layer to convert inputs into a quasi-natural picture space. Other defensive strategies typically focus on adversarial purification or robust feature learning. AutoEncoders reconstruct clean approximations of perturbed inputs, suppressing adversarial noise. MagNet \citep{mengMagNetTwoProngedDefense2017} combines detection and purification AutoEncoders to project adversarial samples back onto the manifold of natural images. D2Defend \citep{yanD2DefendDualDomainBased2021} introduces a dual-domain denoising strategy that leverages bilateral filtering in the spatial domain and wavelet-based shrinkage in the frequency domain to isolate and remove adversarial signals.\\

\end{document}